\documentclass[final,1p,times,authoryear]{elsarticle}

\usepackage[round]{natbib}
\usepackage[pdftex,bookmarks=true,pdfborder={0 0 0}]{hyperref} 
\usepackage[spacing=true,protrusion=true]{microtype}
\usepackage{amsmath,amsfonts,amssymb}
\usepackage{graphicx,subfig,color}
\usepackage{setspace,booktabs,calc,enumitem,placeins}
\usepackage{tikz}
\usetikzlibrary{shapes,positioning,arrows.meta,backgrounds}
\usepackage[absolute,overlay]{textpos}
\newcommand{\nil}{@{}}
\newcommand{\nl}{\newline}
\newcommand{\rbox}[2]{\rotatebox{45}{\parbox{#1\textwidth}{#2}}}

\newcommand{\btab}{
	\center
	\scriptsize
	\bgroup\def\arraystretch{1.2}
	\begin{tabular}{\nil p{0.32\textwidth} @{\hspace{11pt}} l \nil l \nil l \nil l \nil l \nil l \nil l \nil l \nil}
		\toprule
		\bf Paper	& \multicolumn{5}{l}{\hspace{-5pt}\bf Agents} & \multicolumn{3}{l}{\hspace{-5pt}\bf Environment} \\[-5pt]
				& \rbox{0.075}{Stochastic \nl actions?} & \rbox{0.075}{Changing \nl behaviour?} & \rbox{0.075}{Factors \nl known?} & \rbox{0.07}{Independent \nl agents?} & \rbox{0.06}{Common \nl goals?} & \rbox{0.10}{Move order} & \rbox{0.11}{State/action \nl representation} & \rbox{0.09}{State/action \nl observability} \\
		\midrule}
\newcommand{\tabrow}[9]{\citep{#1} & #2 & #3 & #4 & #5 & #6 & #7 & #8 & #9 \\}

\newcommand{\etab}{
		\bottomrule
	\end{tabular}
	\egroup}

\newcommand{\si}{simult.}
\newcommand{\al}{altern.}
\newcommand{\as}{$^{*}$}
\newcommand{\ass}{$^{**}$}
\newcommand{\asss}{$^{***}$}

\newcommand{\proconlist}{\vspace{-3pt}\begin{itemize}[leftmargin=10pt,itemsep=-1pt]}
\newcommand{\itempro}{\renewcommand{\labelitemi}{$+$}\item}
\newcommand{\itemcon}{\renewcommand{\labelitemi}{$-$}\item}
\newcommand{\elist}{\end{itemize}}

\journal{Artificial Intelligence}

\begin{document}


	\begin{frontmatter}

		\title{Autonomous Agents Modelling Other Agents: \\ A Comprehensive Survey and Open Problems}

		\author[ed]{Stefano V. Albrecht}
		\address[ed]{The University of Edinburgh, United Kingdom}
		\author[ut]{Peter Stone}
		\address[ut]{The University of Texas at Austin, United States}

		\begin{abstract}
Much research in artificial intelligence is concerned with the development of autonomous agents that can interact effectively with other agents. An important aspect of such agents is the ability to reason about the behaviours of other agents, by constructing \emph{models} which make predictions about various properties of interest (such as actions, goals, beliefs) of the modelled agents. A variety of modelling approaches now exist which vary widely in their methodology and underlying assumptions, catering to the needs of the different sub-communities within which they were developed and reflecting the different practical uses for which they are intended. The purpose of the present article is to provide a comprehensive survey of the salient modelling methods which can be found in the literature. The article concludes with a discussion of open problems which may form the basis for fruitful future research.
		\end{abstract}


	\end{frontmatter}


	\tableofcontents
	\noindent\hrulefill

	\section{Introduction} \label{sec:intro}

\begin{figure}[t]
	\center
	\includegraphics[height=0.074\textheight]{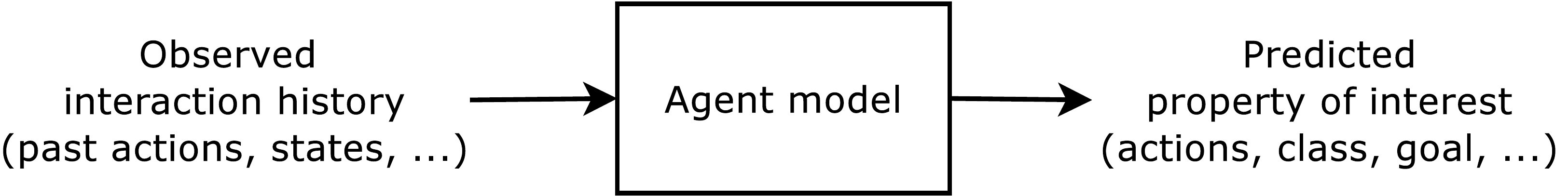}
	\caption{General agent model.}
	\label{fig:model}
\end{figure}

A core area of research in modern artificial intelligence (AI) is the development of autonomous agents that can interact effectively with other agents. An important aspect of such agents is the ability to reason about the behaviours, goals, and beliefs of the other agents. This reasoning takes place by constructing \emph{models} of the other agents. In general, a model is a function which takes as input some portion of the observed interaction history, and returns a prediction of some property of interest regarding the modelled agent (cf. Figure~\ref{fig:model}). The interaction history may contain information such as the past actions that the modelled agent took in various situations. Properties of interest could be the future actions of the modelled agent, what class of behaviour it belongs to (e.g. ``defensive'', ``aggressive''), or its current goals and plans.

An autonomous agent can utilise such a model in different ways, but arguably the most important one is to inform its decision making. For example, if the model makes predictions about the actions of the modelled agent\footnote{We will use the term ``modelling agent'' to refer to the agent which is carrying out the modelling task, and ``modelled agent'' or ``other agent'' to refer to the agent which is being modelled.}, then the modelling agent can incorporate those predictions in its planning procedure to optimise its interaction with the modelled agent. If instead the model makes predictions about the class of behaviour of the modelled agent, then the modelling agent could choose a precomputed strategy which it knows to work well against the predicted class. Besides informing decisions, an agent model can also be used for other purposes. For example, an intelligent tutoring system could use a model of a specific human player in games such as Chess to identify and point out weaknesses in the human's play \citep{iku1996}.

The process of constructing models of other agents, sometimes referred to as \emph{agent modelling} or \emph{opponent modelling},\footnote{Because much of the early work was developed in the context of competitive games such as Chess, the term ``opponent modelling'' was established to refer to the process of modelling other agents, and is still used by many researchers.} often involves some form of learning since the model may be based on information observed from the current interaction and possibly data collected in past interactions. For example, an agent may model another agent's decision making as a deterministic finite automaton and learn the parameters of the automaton (e.g. nodes, edges, labels) during the interaction \citep{cm1998dfa}. Similarly, an agent may attempt to classify the strategy of another agent by using classifiers which were trained with statistical machine learning on data collected from recorded interactions \citep{wm2009}.

Modelling other agents in complex domains is a challenging task. In the above example in which an agent models another agent's behaviour as a finite automaton, the learning task is known to be NP-complete in both the exact and approximate cases \citep{pitt1989,g1978}. Many other modelling techniques exist, each with their own complexity issues. For example, the task of inferring an agent's goals and plans based on complex action hierarchies often faces an exponential growth in plan hypotheses \citep{g2004}. Yet, despite such difficulties, research in modelling other agents continues to push the boundary, in part driven by innovative applications that necessitate effective modelling capabilities in agents. For example, dialogue systems have to understand and disambiguate the intentions and plans of users \citep{gs1986,la1984}; intelligent tutor systems must reason about the knowledge and misconceptions of students to facilitate learning progress \citep{mvgb2000,abcl1990}; autonomous military and security systems must be able to reason about the decision making, beliefs, and goals of adversaries \citep{bkaa2015,jlm2005,t1995rec}; and autonomous vehicles must reason about the behaviours of other vehicles \citep{bls2009}. Beyond such applications of ``narrow AI'', there is also the grand vision of a general AI which is capable of completing tasks, across different domains, that potentially require non-trivial interactions with other agents (including humans). It is evident that such a general AI will require an ability to reason about the goals, beliefs, and decision making of other agents. This is especially true in the absence of coordination and communication protocols, where modelling other agents is a key requirement for effective collaboration \citep{skkr2010,rw2003}.

There is a rich history of research on computational agents that model other agents. Some of the earliest work can be traced back to the beginnings of game theory, in which opponent modelling was studied as a means of computing equilibrium solutions for games. The classical example is ``fictitious play'' \citep{b1951}, in which each player models the other player's strategy as the empirical frequency distribution of their past play. Another example is rational learning \citep{kl1993}, in which players maintain Bayesian beliefs over a space of possible strategies for the other players. In AI research and computational linguistics, methods for recognising the goals and plans of agents \citep{ssg1978} were applied in automated dialogue systems to understand and disambiguate the intentions of users \citep{p1986,la1984}. Adversarial games such as Chess were also an important driver of research in opponent modelling. The dominant solution for such games was based on the ``minimax'' principle, in which agents optimise their decisions against a worst-case, foolproof opponent \citep{cm1983}. However, it was recognised that real players often exhibit limitations in their strategic play, e.g. due to cognitive biases or bounded computation, and that knowledge of such limitations could be exploited to obtain superior results to minimax play \citep{iuhh1994,iuhh1993,cm1993search,rb1983}. In addition to opponent modelling in game playing, early models of recursive reasoning (``I believe that you believe that I believe...'') were formulated \citep[e.g.][]{gdw1991,wb1986}. Since these early works in game theory and AI, the problem of modelling other agents has been an active area of research in many sub-communities, including classic game playing \citep{f2001}, computer Poker \citep{rw2011poker}, automated negotiation \citep{bhhj2016}, simulated robot soccer \citep{1998robocup}, human user modelling \citep{za2001,m1993}, human-robot interaction \citep{lfs2014}, commercial video games \citep{bsl2012}, trust and reputation \citep{rhj2004}, and multiagent learning \citep{sv2000}.

Many different modelling techniques now exist which vary widely in their underlying assumptions and methodology, largely due to the different needs and constraints of the sub-communities within which they were developed. Assumptions may pertain to aspects of the modelled agent, such as whether the agent makes deterministic or stochastic action choices, and whether its behaviour is fixed or may change over time. They may also pertain to aspects of the environment, such as whether the actions of other agents and environment states are observed fully or only partially with possible uncertainty. Current methodologies include learning detailed models of an agent's decision making as well as reasoning about spaces of such models; inferring an agent's goals and plans based on hierarchical action descriptions; recursive reasoning to predict an agent's state of mind and its higher-order beliefs about other agents; and many other approaches. While some articles have surveyed modelling methods specific to one of the aforementioned sub-communities (see Section~\ref{sec:relwork}), there is a gap in the current literature in that there is no unified survey of the principal modelling methods which can be found across the sub-communities. As a result, there has been a missed opportunity to effectively communicate ideas, results, and open problems between these sub-communities.

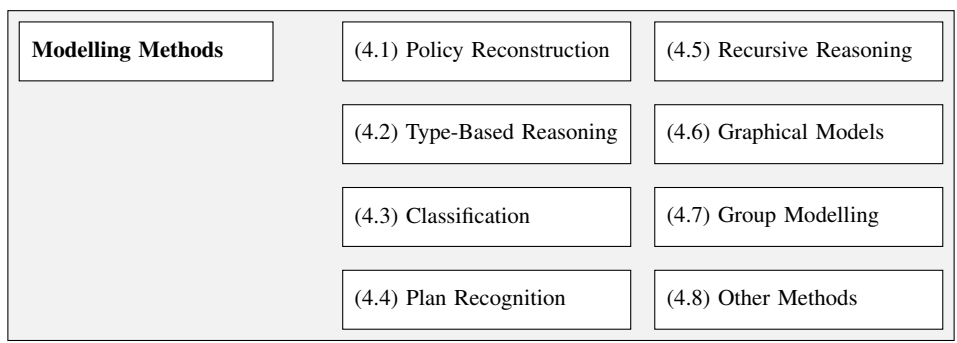
\begin{figure}[t]
	\small
	\center
	\begin{tikzpicture}[background rectangle/.style={draw,fill=gray!10}, show background rectangle, node distance=1em, every node/.style={draw, rectangle, inner sep=5pt, outer sep=0pt, text width=11.5em, minimum height=2.6em, align=left, fill=white, scale=0.95}, every path/.style={-{Latex[length=6pt, width=4pt]}}]
		\node (n0) [text width=10em, outer sep=2em] {\bf Modelling Methods};
		\node (n1) [right=of n0] {(\ref{sec:policyrec}) Policy Reconstruction};
		\node (n2) [below=of n1] {(\ref{sec:typebased}) Type-Based Reasoning};
		\node (n3) [below=of n2] {(\ref{sec:classif}) Classification};
		\node (n4) [below=of n3] {(\ref{sec:planrec}) Plan Recognition};
		\node (n5) [right=of n1] {(\ref{sec:recreas}) Recursive Reasoning};
		\node (n6) [below=of n5] {(\ref{sec:graphmod}) Graphical Models};
		\node (n7) [below=of n6] {(\ref{sec:groupmod}) Group Modelling};
		\node (n8) [below=of n7] {(\ref{sec:other}) Other Methods};
	\end{tikzpicture}
	\caption{Surveyed modelling methods. Brackets show linked section numbers.}
	\label{fig:methods}
\end{figure}

The purpose of the present article is to provide a comprehensive survey of methods which enable autonomous agents to model other agents, and to highlight important open problems in this area. We identify and describe seven salient modelling methods (plus other relevant methods) which are shown in Figure~\ref{fig:methods}. Works were included in the survey if a significant part of the work was concerned with the problem of modelling other agents, which in most cases included the proposal of novel algorithms and/or analysis of and experiments with existing algorithms.

After discussing related surveys in Section~\ref{sec:relwork}, we begin our survey in Section~\ref{sec:assum} with a discussion of the different assumptions that modelling methods may be based on, to help the reader gain an understanding of the applicability and limitations of methods. Section~\ref{sec:methods} then surveys a number of different modelling methods by discussing the general idea underlying each method and surveying the relevant literature. Section~\ref{sec:openprob} concludes with a discussion of open problems which have not been sufficiently addressed in the literature, and which may be fruitful avenues for future research.

	\section{Related Surveys} \label{sec:relwork}

Several articles survey research on opponent modelling for specific domains. \citet{bhhj2016} provide a survey of opponent modelling in bilateral negotiation settings, in which two agents negotiate the values of one or more ``issues'' (e.g. cost, size, and colour of a car) in an exchange. \citet{bsl2012} and \citet{kzc2014} survey methods for player modelling in commercial video games, where the purpose of modelling is to improve the playing strength of game AI as well as player satisfaction. \citet{pd2012} provide an overview of modelling methods used in 2D simulated robot soccer, in which two teams of agents compete in a soccer match. \citet{rw2011poker} survey research in Poker playing agents and dedicate a section to opponent modelling methods. \citet{lfs2014} survey research in safe human-robot interaction and include a section on methods that predict the motions and actions of humans. Several articles survey work in trust and reputation modelling in multiagent systems \citep[e.g.][]{ps2013,yslml2013,rhj2004}. Other surveys of opponent modelling include \citet{hds2005}, \citet{om2005}, and \citet{f2001}. The above articles survey modelling methods for specific domains, and their discussions are centred on the particular properties of interest (e.g. offer preferences, team formation, action timing, human motion, trust levels) and constraints (e.g. limited computational resources, extensive form games of imperfect information, modelling from raw data) in these domains.

Our article is a general survey of the major modelling methods that can be found across the literature, including methods which are not or only sparsely addressed in the above surveys, such as type-based reasoning, plan recognition, recursive reasoning, and graphical models. In contrast, the above surveys primarily focus on specific interaction settings which differ significantly in their rules, dynamics, and assumptions, with many of the surveyed methods being domain-specific. While, ultimately, it is useful to exploit specific domain structure to achieve optimal performance, a focus on domain-specific aspects can make it difficult for researchers unfamiliar with the subject to gain an understanding of the general modelling approaches and, thus, contributes to a fragmentation of the community, as evidenced by the fact that the above surveys have little overlap in terms of cited works. Still, one can identify common ideas in methodology between these communities, such as the use of machine learning methods to ``classify'' other agents and the use of Bayesian beliefs to reason about the relative likelihood of alternative models. Our survey aims to distil the broader context of such methodologies and to provide an overview of the relevant works as well as discuss open problems and avenues for future research, thus documenting the state-of-the-art in agent modelling methods.

In addition to the above surveys, there are also a number of surveys on the topic of multiagent learning \citep{hkbm2017,bthk2015,tw2012,bbd2008,pl2005,adkln2001,sv2000,sw1999}. Multiagent learning\footnote{The 2017 International Joint Conference on Artificial Intelligence held a tutorial on ``Multiagent Learning: Foundations and Recent Trends''. Tutorial slides can be downloaded at: \url{http://www.cs.utexas.edu/~larg/ijcai17_tutorial}} (MAL) is defined as the application of learning to facilitate interaction between multiple agents, where the learning is typically carried out by the individual agents or some central mechanism that has control over the agents. Modelling other agents often involves some form of learning about the other agents and can, thus, be viewed as a part of MAL. However, MAL may also involve other types of learning, such as learning to coordinate without constructing models of other agents \citep[e.g.][]{ar2012,bv2002,hm2001} and learning based on communication. Most of the cited MAL surveys provide some discussion of research on modelling other agents, but due to the broader scope the discussions are necessarily limited. Moreover, some of these surveys are somewhat dated now (albeit still useful), and miss out on much of the more recent progress in modelling methods.

A complicating factor in complex domains such as human-robot interaction, simulated robot soccer, and some commercial games is the fact that agents cannot directly observe the chosen actions of other agents, but must instead infer these (with possible uncertainty) from other observations, such as changes in the environment. The task of identifying actions from raw sensor data and changes in states is referred to as \emph{activity recognition}, and it is itself an active research area that has produced a substantial body of work \citep{sgb2014}. Methods for activity recognition are not covered in our survey. We assume that the modelling agent has some means to identify actions during the interaction, e.g. by using domain-specific heuristics as is often done in the robot soccer domain \citep[e.g.][]{kfcv2002}, training an action classifier using supervised machine learning \citep[e.g.][]{lasb2009}, or reasoning about the probabilities of possible observations \citep[e.g.][]{pg2017}.

	\section{Assumptions in Modelling Methods} \label{sec:assum}

Before surveying the modelling methods, we will discuss some of their possible underlying assumptions. This discussion will be useful for appreciating the applicability and limitations of methods, as well as where some of the current open problems lie. We categorise assumptions into assumptions about the modelled agents and assumptions about the environment within which the agents interact. (For example, in a soccer game, the environment is defined by the soccer field and ball/player positions, and the game dynamics.) 

The following is a list of possible assumptions about the modelled agent. To make this discussion a little more precise, we will use $P(a_j | H)$ to denote the probability with which the modelled agent $j$ chooses action $a_j$ after some history $H = \langle o^1,o^2,...,o^t \rangle$, where $o^\tau$ is an observation at time $\tau$ and $t$ is the current time step. For example, under a fully observable setting, $o^\tau$ may include the environment state at time $\tau$ and the actions of other agents (if any) at time $\tau-1$.

\begin{description}
	\item[Deterministic or stochastic action choices?]  An agent makes deterministic action choices if for every history $H$, $P(a_j | H) = 1$ for some action $a_j$. The more general case are stochastic action choices, in which actions may be chosen with any probabilities.\footnote{In the game theory literature, stochastic actions are often referred to as ``mixed strategies'' \citep[e.g.][]{m1991gt}} Assuming deterministic action choices can greatly simplify the modelling task because we can be sure that the modelled agent will always choose the same action for a given history. This allows us to use deterministic structures such as decision trees and deterministic state automata, for which efficient learning algorithms exist. Besides simplifying the learning of models, assuming deterministic action choices can also simplify the planning of our own agent's actions, because the planning does not have to account for uncertainties in the modelled agent's actions. On the other hand, such an assumption precludes the possibility that the modelled agent may randomise deliberately or that it may make mistakes, as human agents often do. Therefore, modelling methods which allow for stochasticity in action choices can facilitate more robust prediction and planning.

	\item[Fixed or changing behaviour?]  An important question in modelling methods is the degree to which the modelled agent is allowed to change its decision making. The precise meaning of change varies in the literature and also depends on the property of interest that is to be predicted (e.g. actions, class, plan). The basic notion is that the modelled agent has some ability to adapt its decision making based on its past observations. An example of a non-changing (sometimes called ``fixed'', ``stationary'', or ``non-learning'') agent often found in the literature is a simple ``Markovian'' agent which chooses its actions based only on the most recent observation and regardless of what happened before, i.e. $P(a_j | H) = P(a_j | o^t)$. In contrast, an example of an adaptive/learning agent is one which itself tries to learn models of other agents and bases its decisions on these models. Early modelling methods assumed fixed behaviours to avoid the added complexity of tracking and predicting possible changes in behaviours. Today, more methods allow for varying degrees of adaptability in order to allow for greater complexity in modelled agents.

	\item[Decision factors known or unknown?]  Agents often make decisions based on some portion of the history (e.g. the most recent $n$ observations), or based on abstract features which were calculated from the history. An example of an abstract feature is the average number of times a particular action was observed in a specific situation. Given such dependencies on factors, an important question in modelling methods is whether the relevant factors in the modelled agent's decision making are known a priori. Many methods assume that this knowledge is available, or that the relevant factors can in principle be derived from the information available in the observed history. In the worst case, the modelling method can work on the entire history and the hope is that the relevant factors are approximately reconstructed in the modelling process. However, if such a reconstruction is not possible and knowledge of relevant factors is not available, then the predictions of the resulting model can be very unreliable. Some methods attempt to solve this problem by reasoning about a space of possible relevant factors (cf. Section~\ref{sec:policyrec-conditional}).

	\item[Independent or correlated action choices?]  If the modelling agent is interacting with more than one other agent, then a possible question is whether the other agents choose their actions independently from each other. Independence means that the joint probability $P(a_j, a_{j'} | H)$ for agents $j$ and $j'$ can be factored into $P(a_j|H)P(a_{j'}|H)$. Otherwise, the agents are said to have correlated action choices. Many modelling methods assume independent action choices, which allows for the independent construction of models for each agent. Note that independence does not mean that the agents ignore each other, since they may observe each others' past actions in the history $H$. However, if agents are correlated in their action choices, e.g. due to joint plans and communication \citep{sv1999aij,gk1996}, then it may be important for the modelling method to capture such correlations. For applications in which this is the case, such as robot soccer, researchers have developed methods that model entire teams as opposed to individual agents.

	\item[Common or conflicting goals?]  Another possible assumption concerns the agents' goals.\footnote{Assumptions about the goals of agents may also be viewed as assumptions about the environment, since the payoff/reward functions are usually part of the task and environment specification. We view them as assumptions about agents to allow for the more general notion of subjective goals, such as intrinsic rewards \citep{sbc2005}.} A goal may be to reach a specific state in the environment or to optimise a given objective function, such as the payoff/reward functions used in game theory and reinforcement learning. Goals are said to be common if they are identical for all agents. Many modelling methods that attempt to predict an agent's actions are unaffected by the goals of the agents, since such methods primarily work on observed actions (cf. Sections~\ref{sec:policyrec} and \ref{sec:typebased}). However, methods which attempt to predict the intentions and beliefs of other agents can be influenced significantly by assumptions about goals, since an observed action may yield different clues when viewed in the context of common versus conflicting goals. Some modelling methods attempt to learn the payoff functions used by other agents (cf. Section~\ref{sec:policyrec-utility}).
\end{description}

In addition to assumptions about the modelled agent, many methods make assumptions about the environment within which the interaction takes place. Some common assumptions concern the order in which agents choose their actions (simultaneous or alternating moves), and the representation of actions and environment states (discrete, continuous, mixed). However, the most important assumptions usually concern the extent to which agents are able to observe what is happening in the environment. Much of the early work in opponent modelling was developed in idealised settings such as Chess, in which the state of the environment and the agents' chosen actions are fully observable by all agents. The domain of Poker added the problem of partial observability of environment states, since no player can see the private cards of other players. In domains such as human-robot interaction and robot soccer, additional complications are that observations about the environment state may be unreliable (e.g. due to noisy sensors), and that actions may no longer be observed directly by the agents but have to be inferred (with some uncertainty) based on other observations, such as changes in the environment. (For example, a soccer player may infer a passing action between two players based on changes in the position, velocity, and direction of the ball.) Such partial observability can make the modelling task significantly more difficult, since agents can make decisions based on private observations and the modelling method must take such possibilities into account.


	\section{Modelling Methods} \label{sec:methods}

This section provides a comprehensive survey of the salient modelling methods that can be found in the literature (cf. Figure~\ref{fig:methods}). Specifically, we will survey methods of policy reconstruction (Section~\ref{sec:policyrec}), type-based reasoning (Section~\ref{sec:typebased}), classification (Section~\ref{sec:classif}), plan recognition (Section~\ref{sec:planrec}), recursive reasoning (Section~\ref{sec:recreas}), graphical models (Section~\ref{sec:graphmod}), group modelling (Section~\ref{sec:groupmod}), and other relevant methods (Section~\ref{sec:other}). For each modelling method, we provide a table\footnote{See \ref{sec:appendix} for further clarifications on assumption tables.} which lists the assumptions in the surveyed papers, organised according to the dimensions identified in Section~\ref{sec:assum}. Table~\ref{tab:methodsummary} provides a high-level summary of the surveyed modelling methods.

\begin{table}[h]
	\center
	\scriptsize
	\bgroup\def\arraystretch{1.7}
	\begin{tabular}{\nil p{0.24\textwidth} \nil p{0.757\textwidth} \nil}
		\toprule
		\bf Method & \bf Summary \\[4pt]

		Policy reconstruction (\ref{sec:policyrec}) &
		Model predicts action probabilities of modelled agent. Assume specific model structure and learn model parameters based on observed actions.
		\proconlist
			\itempro Can learn arbitrary model (subject to chosen model structure)
			\itempro Models often progressively generated during the interaction
			\itemcon May require many observations to yield useful model
			\itemcon Learning task can be complex (space/time)
		\elist \\

		Type-based reasoning (\ref{sec:typebased}) &
		Model predicts action probabilities of modelled agent. Assume agent has one of several known types and compute relative likelihood of types based on observed actions.
		\proconlist
			\itempro Types can be very general (e.g. blackbox)
			\itempro Can lead to fast adaptation if true type of agent (or a similar type) is in type space
			\itemcon Can lead to wrong predictions if type space is wrong
			\itemcon Beliefs not expressive enough to tell if type space is wrong
		\elist \\

		Classification (\ref{sec:classif}) &
		Model predicts class label (or real number, if regression) for modelled agent. Choose model structure and use machine learning to fit model parameters based on various information sources.
		\proconlist
			\itempro Can learn to predict various kinds of properties
			\itempro Many machine learning algorithms available
			\itemcon Learning may require large amount of data to yield useful model
			\itemcon Model is usually computed before interaction and can be difficult to update during interaction
		\elist \\

		Plan recognition (\ref{sec:planrec}) &
		Model predicts goal and (to some extent) future actions of modelled agent. Algorithms often use hierarchical plan library or domain model.
		\proconlist
			\itempro Knowledge of goal and plan extremely useful for long-term planning
			\itempro Rich plan library can encode complex plans (e.g. with temporal and applicability conditions)
			\itemcon Specifying plan library can be tedious/impractical; may be incomplete
			\itemcon Most methods assume modelled agent is unaware of observer (``keyhole plan recognition'')
		\elist \\

		Recursive reasoning (\ref{sec:recreas}) &
		Model predicts next action of modelled agent. Recursively simulate reasoning of modelled agent (``I think that you think that I think...'').
		\proconlist
			\itempro Account for higher-order beliefs of other agents
			\itemcon Recursion is computationally expensive
			\itemcon Assumes modelled agent is rational
		\elist \\

		Graphical models (\ref{sec:graphmod}) &
		Model predicts action probabilities of modelled agent. Uses graphical model to represent agent's decision process and preferences.
		\proconlist
			\itempro Detailed model of agent's domain conceptualisation (causal beliefs) and preferences
			\itempro Graphical representation can lead to computational improvements
			\itemcon Does not scale efficiently to sequential decision processes
		\elist \\

		Group modelling (\ref{sec:groupmod}) &
		Model predicts joint properties of group of agents (e.g. joint action/goal/plan of group).
		\proconlist
			\itempro Can capture correlations in action choices of group
			\itempro Can exploit group structure to improve efficiency and quality of prediction
			\itemcon Reasoning about agent groups is highly complex due to interdependencies among agents
		\elist \\

		\bottomrule
	\end{tabular}
	\egroup
	\caption{High-level summary of surveyed modelling methods, with an indication of some of their potential strengths ($+$) and limitations ($-$). This summary does not apply to all surveyed papers; many variations exist, and not all potential strengths and limitations are listed in this summary (see main text).}
	\label{tab:methodsummary}
\end{table}

\FloatBarrier

		\subsection{\bf Policy Reconstruction} \label{sec:policyrec}

Policy reconstruction methods generate models which make explicit predictions about an agent's actions, by reconstructing the agent's decision making. Most methods begin with some arbitrary or idealised model and ``fit'' the internals of the model to reflect the agent's observed behaviour. The predictions of such a model can be utilised by a planner to reason about how the modelled agent might react to various courses of actions. For example, Monte-Carlo tree search \citep{bpwl2012} can naturally integrate such models to sample possible interaction trajectories, which are used to find optimal actions with respect to the agent model.

The two central design questions in policy reconstruction methods are (1) what elements of the interaction history should be used to make predictions, and (2) how should these elements be mapped to predictions? The following discussion of methods gradually shifts emphasis from the first question to the second question.

		\subsubsection{Conditional Action Frequencies} \label{sec:policyrec-conditional}

The archetypal example of a policy reconstruction method is ``fictitious play'' \citep{b1951}, in which agents model each other as a probability distribution over their possible actions. The probabilities are ``fitted'' via a maximum-likelihood estimation over the agents' observed actions, which corresponds to simply computing their average frequencies. This simple method has some well-known convergence properties in matrix games \citep{fl1998} and was adopted early in multiagent reinforcement learning \citep{cb1998}. Of course, a single distribution is unable to capture agent behaviours with complex dependencies on the interaction history. The key to making this method more capable is to \emph{condition} the action distribution on elements of the history. For instance, \citet{sa1997} and \citet{bs2007} propose agents that learn the action frequencies of other agents conditioned on the modelling agent's own action, and \citet{dh1998} propose a user model which learns conditional probabilities of user commands based on the user's previous command. More complex methods may condition distributions on more information from the history, such as the $n$ most recent joint actions of all agents \citep{ps2005}.

The difficulty with learning conditional action distributions is that we may not know what elements of the history to use. If we condition distributions on too little or the wrong information from the history, then the learned distributions may not produce reliable predictions. If we condition on too much information, then the learning may be too slow and inefficient. To address this issue, methods have been developed which automate the conditioning. \citet{jbgs2005} propose a method which learns action frequencies for each possible subset of the $n$ most recent elements in the history. To manage the combinatorial explosion of subsets, some subsets are removed if the entropy of their conditional distributions is above some threshold, meaning that their predictions are not certain enough. To make a prediction, the method selects the subset with the lowest entropy for the given history (i.e. most certain prediction). Similarly, \citet{cs2014} describe a method which learns action frequencies conditioned on the most recent $n, n-1, n-2, ...$ observations and plans its own actions using the ``smallest'' conditioning which best predicts the modelled agent's actions, in the sense that it is not too dissimilar to the largest conditioning. Essentially the same method can be used to model agents which condition their choices on abstract feature vectors derived from the history \citep{cs2013}.

The idea of monitoring conditional action frequencies of the modelled agents has also been used in the context of extensive form games with imperfect information, such as Poker \citep{ms2017,gs2011,sbl2005,bdsb2004}. Such games are characterised by the fact that agents may have private information (e.g. cards in own hand) in addition to public information (e.g. cards on the table). Hence, agents make decisions in ``information sets'', which are sets of decision nodes that cannot be distinguished with the available information. The decision making of other agents can be modelled as the observed frequency with which they chose actions in the various information sets. For example, \citet{sbl2005} associate an independent Dirichlet distribution for each information set and update the corresponding distribution after each observed action. Dirichlet distributions are a natural way to model uncertainty over finite probability distributions and can be updated efficiently. Rather than learning such distributions from scratch, it is also possible to initialise the distributions to some reasonable baselines. For example, \citet{gs2011} first compute a Nash equilibrium solution for the game which specifies action distributions for each information set and agent. This solution can be used to initialise the agent models. During play, the distributions in the models are gradually shifted toward the observed action frequencies of the modelled agents, to reflect their true behaviours. The advantage of this method is that the modelling agent can initially plan its actions against a rational (Nash) opponent model, rather than starting with an arbitrary model. \citet{bdsb2004} propose a method which learns action frequencies conditioned on entire action sequences. To generalise observed actions more quickly, the method employs a sequence of increasingly coarse abstractions over action sequences. Moreover, to allow for changing behaviours, the method uses a decay factor such that more recent observations have greater weight in the calculation of action frequencies.

		\subsubsection{Case-Based Reasoning} \label{sec:policyrec-casebased}

A limitation in the above methods is that they may lack a mechanism to extrapolate (or ``generalise'') past observations to previously unseen situations. Abstraction methods such as those used by \citet{bdsb2004} can achieve some level of generalisation by defining equivalence relations over observations. Case-based reasoning \citep[e.g.][]{k2014,v1994,h1986} is a related method which uses similarity functions to relate observations. In essence, this method maintains a set of ``cases'' along with the observed actions of the modelled agent in each encountered case. To extrapolate between cases, a similarity function must be specified which measures how similar two given cases are. For example, in simulated robot soccer, a case may be defined by the state of the soccer field, and the similarity could measure the respective differences of ball and players positions in two given cases. When presented with a new case, the method searches for the most similar known cases and predicts an action as a function of these cases.

\citet{ar2013} propose a method which stores observed cases (defined as environment states) and the observed actions of the modelled agent in each case. When queried with a new case, the method generates a prediction by searching for similar cases and aggregating their predictions based on the relative similarity to the queried case and the recency of the observed actions to allow for changing behaviours. Similar case-based methods for modelling the behaviour of other agents were proposed by \citet{bkaa2015} and \citet{hs2008}. In all of the above methods, a case is represented as a multi-attribute vector and similarity between vectors is measured using domain-specific difference calculations. An interesting question in case-based methods is whether the similarity function can be optimised automatically with respect to the modelled agent \citep{s2005,s2004moo,aln2003}. For example, \citet{s2004moo} proposes a method in which the similarity function is defined as a linear weighting of differences in the attributes of two given cases. The weighting is learned based on the goal of the modelled agent and a ``Goal Dependency Network'' which specifies dependencies between sub-goals and case attributes. Another important question in case-based methods is how to store and retrieve cases efficiently. For example, \citet{dh2004} propose a retrieval method based on tree search, and \citet{bkaa2015} prune cases to reduce the number of the stored cases.

		\subsubsection{Compact Model Representations}

Methods based on frequency distributions and case-based reasoning are general, since the conditioning and cases can be based on any observable information. However, this generality comes at the cost of exponential space complexity. For example, if action distributions of the modelled agent are conditioned on the past $n$ observations which each can assume $m$ possible values (or, equivalently, if a case consists of $n$ different attributes with $m$ possible values), then there are (up to) $m^n$ distributions to be stored. An alternative method is to use more compact model representations such as those found in the machine learning literature. For example, one may attempt to model an agent's decision making as a deterministic finite automaton (DFA) \citep{cm1998dfa,cm1996dfa,mgr1995}. \citet{cm1996dfa} show how such a model can be learned from observed actions. Essentially, each time the method observes a new action, it checks if the current model is consistent with the observation in the sense that it would have predicted the action, given the current state of the DFA. If it is not, the DFA model is modified to account for the new observation, e.g. by adding new nodes and edges between nodes. A useful property of this method is that it searches for the smallest DFA that is consistent with the observations. Other representations that have been used to model agents include decision trees \citep{bskr2013} and artificial neural networks \citep{alphago2016,dbss2000}.

Machine learning methods can also be used to infer missing information from the observed interaction. For example, in robot soccer an agent cannot directly observe what actions other agents took; it only observes (if at all) the changes in the environment as a result of the agents' actions. \citet{lasb2009} propose a method which trains multiple decision/regression tree classifiers on recordings from past plays. One classifier is trained to predict the action that the modelled agent took, given two consecutive environment states. Another classifier is trained to predict the next action that the agent will take, given the current state and past action predicted by the first classifier. Additional classifiers are trained to predict the continuous parameters of the predicted actions. \citet{pg2017} propose to use probabilistic DFAs (PDFAs) to model the stochastic action choices of agents in domains in which neither the state of the environment nor the other agents' actions are observed with certainty. The proposed method uses a Bayesian nonparametric prior over the space of all PDFAs, and updates the prior after new observations to find a model which captures the behaviour of the modelled agent. \citet{ms2017} use an expectation-maximisation algorithm to infer the current information set of the modelled agent in extensive form games.

		\subsubsection{Utility Reconstruction} \label{sec:policyrec-utility}

One characteristic which is shared by all of the above methods is that they do not model the preferences of the modelled agent, which are often expressed as some kind of utility function. However, it can be difficult to generalise the observed actions from the modelled agent if its preferences are unknown. An alternative is to assume that the modelled agent maximises some utility function which is unknown to the modelling agent. This \emph{rationality} assumption allows the modelling agent to reason about the possible utility function of the modelled agent, given its observed actions. Once an estimate of a utility function is obtained, one can predict the actions of the modelled agent by maximising the utility function from the perspective of the modelled agent.

\begin{table}[t]
	\btab
		\tabrow{ms2017}{yes}{no}{yes}{yes}{no}{\al}{discrete}{partial}
		\tabrow{pg2017}{yes}{yes}{yes}{yes}{no}{\si}{discrete}{partial}
		\tabrow{alphago2016}{yes}{no}{yes}{yes}{no}{\al}{discrete}{full}
		\tabrow{bkaa2015}{no}{no}{yes}{yes}{no}{\si}{mixed}{partial}
		\tabrow{cs2014,cs2013}{yes}{no}{no}{yes}{no}{\si}{discrete}{full}
		\tabrow{ar2013}{yes}{yes}{yes}{yes}{no}{\si}{discrete}{full}
		\tabrow{bskr2013}{no}{no}{yes}{yes}{yes}{\si}{discrete}{full}
		\tabrow{gs2011}{yes}{no}{yes}{yes}{no}{\al}{discrete}{partial}
		\tabrow{lasb2009}{no}{no}{yes}{yes}{no}{\si}{mixed}{partial}
		\tabrow{ht2008}{no}{no}{yes}{yes}{no}{\al}{discrete}{full}
		\tabrow{bs2007}{yes}{no}{--}{yes}{no}{\si}{discrete}{full}
		\tabrow{ps2005}{yes}{no}{no}{yes}{no}{\si}{discrete}{full}
		\tabrow{jbgs2005}{yes}{no}{no}{yes}{no}{\si}{discrete}{full}
		\tabrow{sbl2005}{yes}{no}{yes}{yes}{no}{\al}{discrete}{partial}
		\tabrow{s2005,s2004moo}{no}{no}{no}{yes}{no}{\si}{mixed/disc.}{full}
		\tabrow{bdsb2004}{yes}{yes}{yes}{yes}{no}{\al}{discrete}{partial}
		\tabrow{cj2004}{no}{no}{yes}{yes}{no}{\al}{discrete}{full}
		\tabrow{dh2004}{no}{no}{yes}{yes}{no}{\si}{discrete}{full}
		\tabrow{gpmg2004}{yes}{no}{yes}{yes}{no}{\si}{discrete}{full}
		\tabrow{aln2003}{no}{no}{no}{yes}{no}{\si}{mixed/disc.}{full}
		\tabrow{cko2001}{no}{no}{yes}{yes}{no}{\al}{discrete}{full}
		\tabrow{dbss2000}{yes}{no}{yes}{yes}{no}{\al}{discrete}{partial}
		\tabrow{cb1998}{yes}{no}{--}{yes}{yes}{\si}{discrete}{full}
		\tabrow{dh1998}{yes}{no}{--}{yes}{no}{--\as}{discrete}{full}
		\tabrow{sa1997}{yes}{no}{--}{yes}{no}{\si}{discrete}{full}
		\tabrow{cm1998dfa,cm1996dfa}{no}{no}{yes}{yes}{no}{\si}{discrete}{full}
		\tabrow{cm1996search93,cm1993search}{no}{no}{yes}{yes}{no}{\al}{discrete}{full}
		\tabrow{mgr1995}{no}{no}{yes}{yes}{no}{\si}{discrete}{full}
		\tabrow{b1951}{yes}{no}{--}{yes}{no}{\si}{discrete}{full}
	\etab
	\caption{Assumptions in papers for policy reconstruction methods. \as Does not specify move order.}
	\label{tab:policyrec}
\end{table}

Based on this idea, \citet{cm1996search93,cm1993search} consider opponent modelling in extensive form games (e.g. Checkers) and define a model as the search depth and utility function used by the opponent. The utility function is assumed to be a linear combination of features in the game state, and the goal is to learn the weights in the combination. Given a set of examples which consist of game states and the opponent's chosen action in each state, the proposed method learns multiple candidate models (one for each search depth) using hill-climbing search to iteratively improve the weight estimates until no further improvement is possible. The model which best describes the opponent's moves is then used in the search routine of the modelling agent. \citet{cko2001} consider a similar setting and assume that the modelled agent's utility function is a linear weighting of ``subutilities''. Here, the weighting is known and the goal is to learn the subutilities. Given observed play trajectories, the proposed method generates linear constraints on the space of possible utility functions, similar to methods of inverse reinforcement learning \citep{nr2000}. To select a utility function from the space of possible functions, the authors propose to use a Bayesian prior which is conditioned on observed actions, and the resulting posterior is used to sample a utility function. \citet{gpmg2004} consider single-shot normal-form games and model a human player's utilities as a linear combination of social factors such as social welfare and fairness. Data is collected from human play and utility weight profiles are learned using expectation-maximisation and gradient ascent algorithms. A prior distribution over the different profiles is used to compute expected payoffs for actions.

Learning the utility function, or preferences, of other agents is also a major line of research in automated negotiation agents (see \citet{bhhj2016} for a detailed description of many domain-specific methods). For instance, \citet{ht2008} consider a bilateral multi-issue negotiation and define utility functions as weighted sums of issue evaluation functions. To learn the weights and evaluation functions ascribed by the opponent to each issue, the authors discretise the space of possible weights and evaluation functions by assuming special functional forms. This results in a finite hypothesis space of utility functions over which a Bayesian prior is defined and updated after new bids are received. The resulting posterior can be used to estimate the opponent's utility function. \citet{cj2004} also consider linearly additive utility functions and learn the weights using kernel density estimation. (See also Section~\ref{sec:graphmod} for utility reconstruction methods in graphical models.)


		\subsection{\bf Type-Based Reasoning} \label{sec:typebased}

Learning new models from scratch via policy reconstruction can be a slow process, since many observations may be needed before the modelling process yields a useful model. This can be a problem in applications in which an agent does not have the time or opportunity to collect many observations about another agent. In such cases, it is useful if the agent is able to reuse models learned in previous interactions with other agents, such that it only needs to find the model which most closely resembles the observed behaviour of the modelled agent in the current interaction. In fact, there may be cases in which we know a priori that the modelled agent has one of several known behaviours, and we can provide specifications of those behaviours to the modelling agent.

Based on the above intuition, type-based reasoning methods assume that the modelled agent has one of several known \emph{types}. Each type is a complete specification (a model) of the agent's behaviour, taking as input the observed interaction history and assigning probabilities to the actions available to the modelled agent. Types may be obtained in different ways: they may be specified manually by a domain expert; they may have been learned in previous interactions or generated from a corpus of historical data \citep[e.g.][]{bskr2013}; or they may be hypothesised automatically from the domain and task to be completed \citep[e.g.][]{acr2015}. Given a specification of possible types, type-based reasoning begins with a prior belief which specifies the expected probabilities of types before any actions are observed. During the interaction, each time a new action is observed, the belief is updated according to the probability with which the types predicted the observed action. The modelling agent can then use the updated belief and the types in a planning procedure to compute optimal actions with respect to the types and belief. A useful property of this method is that, if the true type of the modelled agent (or a sufficiently similar type) is in the set of considered types, then the beliefs can often point to this type after only a few observations, leading to fast adaptation. Moreover, since types are essentially blackbox mappings, they can encapsulate policy reconstruction methods to learn new types during the interaction \citep{ar2013,bsk2011}.

Type-based reasoning was first studied by game theorists, who considered games in which all players maintain beliefs about the possible types of the other players \citep{h1967}. The principal questions studied in this context are the degree to which players can learn to make correct predictions through repeated interactions, and whether the interaction process converges to solutions such as Nash equilibria \citep{n1950}. A well-known result by \citet{kl1993} states that, under a certain ``absolute continuity'' assumption regarding players' beliefs, their prediction of future play will get arbitrarily close to the true future play and convergence to Nash equilibrium emerges. (The assumption states that every event with true positive probability is assigned positive probability under the players' beliefs.) Subsequent works studied the impact of prior beliefs on equilibrium convergence and showed that if players have different prior beliefs, their play may converge to a subjective equilibrium which is not a Nash equilibrium \citep{dfl2004,n1998}. Lastly, for certain games and conditions, there are results which show that players cannot simultaneously have correct beliefs and play optimally with respect to their beliefs \citep{n2005,fy2001}.

In AI research, type-based reasoning\footnote{The 2016 AAAI Conference on Artificial Intelligence held a tutorial on ``Type-Based Methods for Interaction in Multiagent Systems''. Tutorial slides can be downloaded at: \url{http://thinc.cs.uga.edu/tutorials/aaai-16.html}} found popularity in problems of multiagent interaction without prior coordination \citep{als2016mipc,skkr2010}, in which the controlled agent interacts with other agents whose behaviours are initially unknown. \citet{acr2016aij} provide a concise and compact definition of a type-based reasoning method via a recursive combination of the Bayes-Nash equilibrium \citep{h1968a} and Bellman optimality equation \citep{b1957}. This combination results in a tree of all possible interaction trajectories as well as their predicted probabilities and payoffs, where the probabilities take into account changes in beliefs along the trajectories. The authors define different belief formulations and analyse their convergence properties \citep{ar2014}. They also show empirically that prior beliefs can have a significant long-term impact on payoff maximisation, and that they can be computed automatically with consistent performance effects \citep{acr2015}. \citet{bsk2011} modify the sampling-based planner UCT \citep{ks2006} such that each rollout in UCT samples a type for each other agent based the current belief over types. The algorithm is evaluated in the ``pursuit'' grid-world domain where it could perform well even if the true types of other agents were not in the set of considered types, so long as sufficiently similar types were known. In subsequent work, \citet{bskr2013} show how transfer learning can be used to adapt decision-tree types learned in previous interactions. \citet{rw2003} propose a method which dynamically learns up to a certain number of types which are represented as deterministic finite automata. When interacting with a new agent, the method finds the closest known type or adds a new type for future reference. Optimal actions against a type are computed using reinforcement learning methods such as Q-learning \citep{wd1992}. \citet{tea2002} propose a ``multi-module'' reinforcement learning method where each module corresponds to a possible agent type and a ``gating signal'' is used to determine how closely each module matches the current agent. Type-based reasoning has also been studied under partial-observability conditions. In Interactive POMDPs \citep{gd2005}, agents have possible uncertainty about the state of the environment, the types of other agents, and their chosen actions. (We defer a more detailed discussion of this model to Section~\ref{sec:recreas}).

The above methods all use Bayes' law or some modification thereof to determine the relative likelihood of types, given the observed actions of the modelled agent. An alternative to Bayes' law is to use machine learning methods such as artificial neural networks, which can learn to predict ``mixtures'' of types (represented as weight vectors) given the observed actions. For example, \citet{lcm2007} propose a method which consists of two neural networks: one network is trained to predict a mixture of types, taking as input the observed actions of the modelled agent; another network is trained to make decisions by assigning probabilities to available actions, taking as input the observed actions and the predicted mixture from the first network. Similarly, \citet{hbkd2016} train a ``gating network'' which combines the predicted Q-values of several ``expert networks'' corresponding to different agent types.

Most type-based reasoning methods use discrete (usually finite) type spaces, where each type is a different decision function. Even inherently continuous hypothesis spaces can be discretised to obtain discrete type spaces \citep[e.g.][]{ht2008}. However, one may also reason directly about continuous type spaces: essentially, we now have a single decision function which has some number of continuous parameters, and the beliefs quantify the relative likelihood of parameter values. A specific parameter setting can then be viewed as one type. For example, \citet{sbl2005} maintain Gaussian beliefs over the continuous parameters of a specified player function for Poker (cf. Table~1 in their paper). It is also possible to combine discrete and continuous type spaces. \citet{as2017} propose a method which reasons simultaneously about both the relative likelihood of a finite set of types \emph{and} the values of any bounded continuous parameters within these types. The method begins with an initial parameter estimate for each discrete type. After new actions are observed, a subset of the types is selected and their parameter estimates updated using methods such as approximate Bayesian updating and exact global optimisation \citep{hpt2000}.

\begin{table}[t]
	\btab
		\tabrow{as2017}{yes}{yes}{yes}{yes}{no}{\si}{discrete}{full}
		\tabrow{sssd2016}{no}{no}{yes}{yes}{no}{\si}{continuous}{full}
		\tabrow{hbkd2016}{yes}{no\ass}{yes}{no}{no}{\si}{mixed}{full}
		\tabrow{acr2016aij,acr2015}{yes}{yes}{yes}{yes}{no}{\si}{discrete}{full}
		\tabrow{ar2014,ar2013}{yes}{yes}{yes}{yes}{no}{\si}{discrete}{full}
		\tabrow{bskr2013,bsk2011}{yes}{no}{yes}{yes}{yes}{\si}{discrete}{full}
		\tabrow{lcm2007}{yes}{no}{yes}{yes}{no}{\al}{discrete}{partial/full}
		\tabrow{gd2005}{yes}{yes}{yes}{yes}{no}{\si}{discrete}{partial}
		\tabrow{n2005}{yes}{yes}{yes}{yes}{no}{\si}{discrete}{full}
		\tabrow{sbl2005}{yes}{no}{yes}{yes}{no}{\al}{discrete}{partial/full}
		\tabrow{dfl2004}{yes}{yes}{yes}{yes}{no}{\si}{discrete}{full}
		\tabrow{cb2003}{yes}{no}{yes}{yes}{no}{\si}{discrete}{full}
		\tabrow{rw2003}{no}{no}{yes}{yes}{no}{\si}{discrete}{full}
		\tabrow{tea2002}{yes}{partial\as}{yes}{yes}{no}{\si}{mixed}{partial}
		\tabrow{fy2001}{yes}{yes}{yes}{yes}{no}{\si}{discrete}{full}
		\tabrow{cm1999}{no}{no}{yes}{yes}{no}{\si}{discrete}{full}
		\tabrow{n1998}{yes}{yes}{yes}{yes}{no}{\si}{discrete}{full}
		\tabrow{kl1993}{yes}{yes}{yes}{yes}{no}{\si}{discrete}{full}
	\etab
	\caption{Assumptions in papers for type-based reasoning methods. \as Types are Markov (non-changing) but modelled agent is assumed to change between types periodically. \ass Modelled agent may change types between episodes but not during episode.}
	\label{tab:typebased}
\end{table}

An interesting aspect of type-based reasoning is the possibility of deliberately choosing actions to elicit information about an agent's type. While it is possible to use schemes such as occasional randomisation in action selection, such schemes ignore the risk that the exploratory actions may influence the modelled agent in unintended ways \citep{cm1999}. In this regard, type-based reasoning can naturally integrate a decision-theoretic ``value of information'' \citep{h1966} into the evaluation of actions. For example, the methods proposed by \citet{cm1999} and \citet{acr2016aij} recursively take into account the potential information that actions may reveal about the type of the modelled agent and how this in turn may affect the future interaction. \citet{cb2003} propose a ``myopic'' approximation of this kind of reasoning which considers only one recursion of belief change, after which beliefs are held constant for the evaluation of actions. \citet{sssd2016} use a form of model predictive control to optimise a heuristic tradeoff between minimising uncertainty in the modelled agent's type and maximising a given reward function. In the related context of goal recognition (cf. Section~\ref{sec:planrec}), the ``Proactive Execution Module'' of \citet{sww2007} incorporates several criteria in the selection of actions, including uncertainty minimisation, expected success, and minimising risk values assigned to actions.

		\subsection{\bf Classification} \label{sec:classif}

While policy reconstruction (Section~\ref{sec:policyrec}) and type-based reasoning (Section~\ref{sec:typebased}) attempt to predict the future actions of the modelled agent, there may be other properties or quantities of interest which an agent model could predict. For example, an agent model may make predictions about more abstract properties such as whether the play style of the modelled agent is ``aggressive'' or ``defensive'' \citep[e.g.][]{sbp2007}, or it may predict quantities such as the expected times at which the modelled agent will take certain actions \citep[e.g.][]{wm2009}. The former task of assigning one of a finite number of labels is referred to as \emph{classification}, whereas the latter task of predicting continuous values is referred to as \emph{regression}. There are different ways in which such predictions can be utilised by a modelling agent. For instance, an assigned class label can be naturally incorporated into the decision procedure of the modelling agent using if-then-else rules or decision trees. Alternatively, given a class label, the agent may employ a precomputed strategy which is expected to be effective against that particular class label.

Classification methods\footnote{We focus on classification methods since many of the surveyed papers in this section are in this category. Note also that regression problems can be transformed into classification problems via a finite discretisation of values, albeit with an exponential growth of class labels if multiple regression variables are jointly discretised.} produce models which assign class labels to the modelled agent (e.g. ``play-style = aggressive'') based on information from the observed interaction. Similarly to policy reconstruction methods, there are two central design questions in classification methods: (1) what observations from the interaction should be used and how should they be represented to facilitate the classification, and (2) how should the classification be performed given the data representation? The second question often includes a learning phase which is carried out prior to the current interaction, using data collected from past interactions.

Several classification methods have been proposed to model players in complex strategy games. \citet{wm2009} propose methods to predict a player's strategy and build times in the game Starcraft. The models are trained on collected replay data from expert human players. Each replay is tagged as one of six strategies and transformed into a feature vector which contains the initial build times for the various unit types in the game. A number of machine learning algorithms (e.g. decision trees, nearest neighbours) are tested on the data and the results show that the learned models can successfully predict player strategies and build times. Using the same collected replay data, \citet{sb2011} propose methods to classify the opening strategy of Starcraft players from a finite set of strategies, using expectation-maximisation and k-means algorithms. \citet{sbp2007} propose domain-specific classifiers to predict the play style (e.g. ``aggressive'', ``defensive'') of players in the game Spring. To account for possible changes in play style, the model prioritises recent observations over past observations. \citet{st2010} use support vector machines (SVMs) \citep{cv1995svm} to predict the ``preferences'' of players in the game Civilization IV. Each player is characterised by integer-valued preferences in areas such as military, cultural, and scientific development. Training data are generated by pitting predefined AI players with different preference settings against each other. The collected data consist of game states which are transformed into feature vectors with attributes such as the number of cities and units. Using the data, one SVM classifier is trained for each preference. \citet{lsma2009} use SVMs to classify the defensive play of opponent teams in the football game Rush 2008. The game specifies finite sets of  team formations and plays for offense and defense. Using game data generated from all combinations of these team formations and plays, a series of multi-label SVM classifiers is trained corresponding to increasing lengths in observation sequences. \citet{ss2007} consider turn-based strategy games such as Dungeons \& Dragons and train SVMs to classify players into a finite set of roles (e.g. ``scout'', ``medic'') using simulated game data for the various roles.

\begin{table}[t]
	\btab
		\tabrow{sb2011}{yes}{no}{--\as}{no}{no}{\si}{mixed}{partial}
		\tabrow{bdfe2010}{no}{no}{--\as}{no}{no}{\si}{mixed}{partial}
		\tabrow{ials2010}{yes}{yes}{--\as}{yes}{no}{\si}{mixed}{partial}
		\tabrow{st2010}{yes}{no}{--\as}{no}{no}{\si}{mixed}{partial}
		\tabrow{lsma2009}{no}{no}{yes}{no}{no}{\si}{mixed}{full}
		\tabrow{wm2009}{yes}{no}{--\as}{no}{no}{\si}{mixed}{partial}
		\tabrow{ilsk2008}{yes}{no}{--\as}{no}{no}{\si}{mixed}{partial}
		\tabrow{sbp2007}{yes}{yes}{--\as}{no}{no}{\si}{mixed}{partial}
		\tabrow{ss2007}{no}{no}{yes}{yes}{no}{\si}{discrete}{full}
		\tabrow{hjs2006}{yes}{no}{--\as}{--\asss}{no}{--\asss}{mixed}{partial}
		\tabrow{s2004}{yes}{no}{yes\ass}{no}{no}{\si}{mixed}{partial}
		\tabrow{mmh2002}{yes}{no}{--\as}{yes}{no}{\si}{discrete}{full}
		\tabrow{vw2002}{yes}{no}{--\as}{yes}{no}{\si}{mixed}{partial}
		\tabrow{ss2001}{yes}{no}{--\as}{--\asss}{no}{--\asss}{mixed}{partial}
		\tabrow{ah2000}{yes}{no}{--\as}{--\asss}{no}{--\asss}{--\asss}{--\asss}
		\tabrow{rv2000}{yes}{no}{--\as}{no}{no}{\si}{mixed}{partial}
		\tabrow{sfr2000}{yes}{no}{--\as}{yes}{no}{\si}{discrete}{full}
	\etab
	\caption{Assumptions in papers for classification methods. \as This assumption does not apply here since the goal is not to predict the actions of agents. \ass Method is in principle based on action prediction and requires specification of decision factors (state descriptions). \asss Not specified.}
	\label{tab:classif}
\end{table}

Another complex domain in which classification methods have been studied is simulated robot soccer. Two notable differences to the above methods are that the models now predict the identities of players or entire teams, and the (partial) use of symbolic methods in addition to statistical machine learning methods. \citet{s2004} proposes the ``Feature-Based Declarative'' classification method. Therein, each model consists of a number of features which are defined as pairs of logical state descriptions and the actions of one or more opponent players expected to be seen in the described states. Compactness of models is achieved by limiting models to features which are highly distinctive (relative to other models) and stable, meaning that they occur frequently for the model. Given an observation of the game, consisting of the game state and player actions, different symbolic approaches and a Bayesian approach can be used to match features to observations. A successful match to the features of a model means that the opponent has been identified. \citet{bdfe2010} propose a relational procedure which works on temporal sequences of game events for a given team. Each sequence consists of high-level actions such as passing and dribbling, which in turn consist of low-level (primitive) actions such as kicking and turning. Inductive logic programming \citep{m1991} is used to automatically select a feature representation from these sequences. Given the feature vectors, the method uses a k-nearest neighbour algorithm with a specified distance function between feature vectors to classify teams. Similarly, \citet{ilsk2008} extract symbolic sequences of game events from which subsequences of a certain length are extracted and their frequencies represented in a ``trie'' structure \citep{f1960}, which is compared to known models using statistical hypothesis testing. This approach has been extended to allow for evolving agent behaviours, essentially by adding new models when the existing ones are found to be insufficient \citep{ials2010}. Other methods proposed for simulated robot soccer include \citet{rv2000}, who classify teams based on a grid discretisation of the playing field which is used to count the occurrence of certain events (such as ball/player positions and pass/dribble events) in specific geographic areas, and \citet{vw2002} who learn decision trees to classify the behaviour of the goal keeper (e.g. ``leaving goal'', ``returning to goal'') and the passing behaviour of opponent players.

Trust and reputation in multiagent systems is an area of research which uses classification and regression methods to model the trustworthiness of agents (see \citet{ps2013}, \citet{yslml2013}, and \citet{rhj2004} for useful surveys). One definition of trust is the expectation with which an agent will realise its terms of a contract in a given context \citep[many other definitions exist, e.g.][]{d2000}. Trust can be based on a multitude of information, including own experiences from interactions with the modelled agent, communicated experiences from other agents in the system, as well as the roles of the modelled agent and its social relations to other agents. For example, \citet{ah2000} classify agents as very trustworthy, trustworthy, untrustworthy, or very untrustworthy based on direct experiences and reported experiences about agents. Many other proposed methods quantify trust as a continuous value which aggregates various information sources using relative importance weights, confidence values, time discounting, etc. \citep[e.g.][]{hjs2006,mmh2002,ss2001,sfr2000}. Such qualitative or quantitative predictions of trust levels can be used by the modelling agent to tailor its interaction with the modelled agent, and, importantly, trust levels can be used to decide which agents to interact with in the first place.

		\subsection{\bf Plan Recognition} \label{sec:planrec}

Plan recognition is the task of identifying the possible goals and plans of an agent, based on the agent's observed actions \citep{c2001}. The focus is on predicting the intended end-product (goal) of the actions that have been observed so far, as well as the sequence of steps (plan) with which the agent intends to achieve its goal.\footnote{``Goal recognition design'' is a closely related problem in which the goal is to modify the environment such that any agent acting in it reveals its goal as early as possible \citep{why2017,kgk2016,kgk2015,kgk2014}.} Knowledge of the goals and plans of other agents can be extremely useful in interactions with them. For example, an adaptive user interface may suggest certain actions and display other relevant information if it knows what the human user intends to accomplish \citep{oh2011,m1993}, and an intrusion detection system may take certain counter measures if it detects the goals and plan of an attacker \citep{gg2001}.

Many plan recognition methods employ a \emph{plan library} which describes the possible plans and goals that the observed agent may pursue. The representation of plans is a key element in plan recognition methods, and many methods use a hierarchical\footnote{Two examples of hierarchical plan libraries are the network security domain of \citet{gg2009} and the pasta-making domain of \citet{ka1986}.} representation in which ``top-level'' goals are decomposed into sub-plans which may be further decomposable. The leaves in this plan hierarchy are the primitive (non-decomposable) actions that can potentially be observed. Plan libraries may also include additional rules such as temporal orderings between the steps in plans, and preconditions on the environment state which must hold in order to perform certain plan steps. Given such a plan library and a set of observed actions, the plan recognition task is to generate possible plan hypotheses that respect the rules of the plan library and explain (i.e. contain) all observed actions. If multiple plan hypotheses exist that explain the observed actions, they may be distinguished by additional factors such as how plausible or probable they are.

Plan recognition differs from policy reconstruction (Section~\ref{sec:policyrec}) and type-based reasoning (Section~\ref{sec:typebased}) in that the latter predict actions for given situations, but they do not predict the intended end-product of these actions, such as that the modelled agent seeks to reach a certain goal state in the environment. On the other hand, while plan recognition can also be used to predict future \emph{actions}, the resulting predictions are often less precise than predictions of models produced by policy reconstruction and type-based reasoning (with some notable exceptions, e.g. \citet{bvw2002}). For example, plans often specify a partial temporal order of actions, such as that some actions have to occur before some other actions. While this flexibility is useful for planning, it leaves open the precise order and probability of actions in a plan execution. Hence, a plan may predict a set of possible actions but not necessarily which action will be taken next.

Plan recognition methods are sometimes categorised into ``keyhole'' and ``intended'' methods \citep{cpa1981}. The difference is in whether the modelled agent is assumed to be aware of the modelling agent. The vast majority of current methods are designed for keyhole plan recognition, in which the modelled agent is assumed to be unaware of the modelling agent.

			\subsubsection{Plan Recognition in Hierarchical Plan Libraries} \label{sec:planrec-hierarch}

\citet{ka1986} propose a symbolic theory of plan recognition in which plans are represented using complex hierarchical actions that decompose into other complex and primitive actions. This results in a graph representation in which edges denote plan decomposition, and root nodes in the graph correspond to ``top-level plans'' which can be interpreted as goals. The recognition problem is then framed as a problem of graph covering given the observed (primitive) actions, which the authors formulate using the concept of circumscription \citep{m1980}. \citet{tr1995} use a hierarchical plan hierarchy in which plan steps are conditioned on environment states. The proposed method commits early to a single plan hypothesis and evaluates new observations in the context of this hypothesis. If the current plan hypothesis is inconsistent with new observations, the method attempts to repair the hypothesis via limited backtracking in the plan hierarchy.  \citet{ak2005} represent the plan library as a directed acyclic graph which specifies decomposition, temporal orderings, and applicability conditions of plan steps. The plan recognition is carried out via a ``lazy'' procedure which time-stamps complete paths in the plan graph that match new observations and respect the temporal orderings and applicability conditions. A complete set of plan hypotheses can then be extracted when needed (hence ``lazy''). Several extensions to this method have been proposed: one which allows for action duration, interleaved plan execution, and missing observations \citep{akz2005}; an extension to rank plan hypotheses by their expected utility to the modelling agent \citep{ak2007}; and an extension which incorporates timing constraints on the plan recognition task \citep{fmbv2014}.

\citet{cg1993} frame plan recognition as a problem of probabilistic inference in Bayesian networks \citep{p1988}. The plan library is represented as a set of decomposable actions, based on which a set of Bayesian networks can be constructed. The root of each network corresponds to a high-level plan for which prior probabilities must be specified, and the child nodes correspond to plan decomposition. The ``belief'' in this plan hypothesis is expressed by the probability that the value of the root node is true, which can be computed using standard inference algorithms \citep{p1988}. \citet{bvw2002} represent plans as a $K$-depth hierarchy of abstract policies, where a policy at depth $k$ selects a policy at depth $k-1$, and policies at depth $k=0$ are the primitive actions. A notable difference from other formulations is that the policies are defined over environment states, which is similar to models learned in policy reconstruction (Section~\ref{sec:policyrec}) and type-based reasoning (Section~\ref{sec:typebased}). The authors show how the recognition process can be framed using dynamic Bayesian networks and they perform inference using the Rao-Blackwellised particle filter \citep{dfmr2000}. A related method is based on probabilistic state-dependent grammars which allow the plan production rules to depend on state information \citep{pw2000}. \citet{gg2009} represent plans based on AND/OR trees, in which AND children are required steps in plans with possible temporal constraints and OR children are alternative (choice) steps in plans of which one must be performed. Their method uses a generative model of plan execution which specifies probabilities for how an agent decides on a particular plan and how the steps in the plan are executed. This plan execution model can be simulated and the authors show how the model can be used to infer plans based on observations.

			\subsubsection{Plan Recognition by Planning in Domain Models} \label{sec:planrec-planning}

Two potential drawbacks of using plan libraries are that their specification can be tedious, and that they may be incomplete (i.e. the observed agent may use a plan that cannot be constructed with the plan library). \citet{rg2009} propose an alternative formulation of plan recognition as a problem of planning in a domain model which is specified in the STRIPS planning language \citep{fn1971}. Given a set of possible goals, the idea is that the potential goals of the observed agent are those goals for which the optimal plans that achieve the goals contain the observed actions in the order in which they were observed. This idea assumes that the modelled agent is ``rational'' in that it only executes optimal plans with respect to a known cost definition (similar to methods of utility reconstruction; cf. Section~\ref{sec:policyrec-utility}). The authors show how existing exact and approximate planning methods can be adopted to compute this set of goals, essentially by solving the planning problem for the modelled agent such that the solution is consistent with the observed actions. This work is subsequently extended to compute Bayesian probabilities over plan hypotheses \citep{rg2010}. Each possible goal now has a specified prior probability, and the likelihood of the observed actions given a goal is defined as the cost difference between the plan that optimally achieves the goal and the plan that optimally achieves the goal and is consistent with the observed actions. This likelihood definition encodes the assumption that an agent is more likely to pursue optimal plans than suboptimal ones. (See also the work of \citet{sru2016} for an alternative probabilistic extension which allows for unreliable observations, and the work of \citet{vk2017} for a heuristic extension that works with continuous domains.) \citet{bst2009,bts2005} propose a very similar idea to \citet{rg2009} but formulate it within Markov decision processes (MDPs) \citep{b1957}. Since MDPs allow for stochasticity in state transitions and action choices, any optimal policy for an MDP that achieves a specific goal induces a likelihood of the observed actions given the goal, which can be used to compute Bayesian posteriors over the alternative goals. Similar goal recognition methods using MDPs were proposed by \citet{capir2011} and \citet{ft2010}. In subsequent work, both \citet{bst2011} and \citet{rh2011} propose planning-based methods to infer the goals (and beliefs) of an agent in partially observable MDPs \citep{klc1998}. \citet{le1995} and \citet{h2001,h2000} propose symbolic graph-based methods for domains specified in extensions of the STRIPS language. Both methods construct graph structures based on the domain model and observed actions, and utilise this structure to find a subset of goals which are consistent with the observed actions.

\begin{table}[t]
	\btab
		\tabrow{vk2017}{yes}{no}{yes}{yes}{no}{--\ass}{continuous}{full}
		\tabrow{sru2016}{yes}{no}{yes}{yes}{no}{--\ass}{discrete}{partial}
		\tabrow{tzk2016}{yes}{yes}{--\as}{yes}{no}{--\ass}{--\as/disc.}{partial}
		\tabrow{fmbv2014}{no}{no}{yes}{yes}{no}{--\ass}{mixed/disc.}{full}
		\tabrow{bst2011}{yes}{yes}{yes}{yes}{no}{--\ass}{discrete}{partial}
		\tabrow{rh2011}{yes}{no}{yes}{yes}{no}{--\ass}{discrete}{partial}
		\tabrow{capir2011}{yes}{no}{yes}{yes}{yes}{\si}{discrete}{full}
		\tabrow{ft2010}{yes}{no}{yes}{yes}{yes}{\si}{discrete}{full}
		\tabrow{rg2010}{yes}{no}{yes}{yes}{no}{--\ass}{discrete}{full}
		\tabrow{g2010}{yes}{no}{yes}{yes}{no}{\si}{discrete}{full}
		\tabrow{gg2009}{yes}{no}{--\as}{yes}{no}{--\ass}{discrete}{partial}
		\tabrow{rg2009}{no}{no}{yes}{yes}{no}{--\ass}{discrete}{full}
		\tabrow{bst2009}{yes}{yes}{yes}{yes}{no}{--\ass}{discrete}{full}
		\tabrow{ak2007}{yes}{no}{yes}{yes}{no}{--\ass}{mixed/disc.}{partial}
		\tabrow{ba2006}{yes}{no}{--\as}{yes}{no}{--\ass}{discrete}{full}
		\tabrow{ak2005}{no}{no}{yes}{yes}{no}{--\ass}{mixed/disc.}{full}
		\tabrow{akz2005}{no}{no}{yes}{yes}{no}{--\ass}{mixed/disc.}{partial}
		\tabrow{bts2005}{yes}{yes}{yes}{yes}{no}{--\ass}{discrete}{full}
		\tabrow{ba2004,ba2003}{yes}{no}{--\as}{yes}{no}{--\ass}{discrete}{full}
		\tabrow{fc2003}{no}{yes}{yes}{yes}{no}{--\ass}{discrete}{full}
		\tabrow{kc2003}{no}{yes}{yes}{yes}{no}{--\ass}{discrete}{full}
		\tabrow{bvw2002}{yes}{no}{yes}{yes}{no}{--\ass}{discrete}{partial}
		\tabrow{h2001,h2000}{no}{no}{yes}{yes}{no}{--\ass}{discrete}{full}
		\tabrow{pw2000}{yes}{no}{yes}{yes}{no}{--\ass}{discrete}{partial}
		\tabrow{azn1998,aznb1997}{yes}{yes}{yes}{yes}{no}{--\ass}{discrete}{partial}
		\tabrow{le1995}{no}{no}{yes}{yes}{no}{--\ass}{discrete}{full}
		\tabrow{tr1995}{no}{no}{yes}{yes}{no}{--\ass}{mixed/disc.}{full}
		\tabrow{bcdk1994}{yes}{no}{yes}{yes}{no}{--\ass}{mixed}{partial}
		\tabrow{cg1993}{no}{no}{--\as}{yes}{no}{--\ass}{--\as/disc.}{full}
		\tabrow{ka1986}{no}{no}{--\as}{yes}{no}{--\ass}{--\as/disc.}{full}
	\etab
	\caption{Assumptions in papers for plan recognition methods. \as Does not model environment states. \ass Does not define move order between agents.}
	\label{tab:planrec}
\end{table}

			\subsubsection{Plan Recognition by Similarity to Past Plans}

Plan hypotheses may also be generated based on similarity to past observed plans. This idea was explored in the context of case-based reasoning methods for plan recognition \citep{kc2003,fc2003,bcdk1994}. For example, \citet{kc2003} represent a plan as a sequence of environment states and actions in each state. Given the current state, a history of observed actions, and a case base consisting of previously observed plans, the recognition task is to retrieve plans from the case base which are similar to the current situation. One way to define similarity is by using state abstractions whereby states that share certain properties are grouped together. A useful property of this approach is that the plan library (case base) does not need to be fully specified ahead of time and can be expanded after new plans have been observed. (See also Section~\ref{sec:policyrec-casebased} for case-based reasoning methods for policy reconstruction.) \citet{tzk2016} formulate plan recognition as a problem of sentence completion in natural language processing. A sentence (plan) is a sequence of words (actions), and the corpus (plan library) consists of previously seen sentences. Based on the corpus, natural language processing methods are used to learn probability distributions for how words may surround other words. An incomplete sentence (plan) can then be completed by filling the missing words such that the overall probability of the resulting sentence is maximised. (See also \citet{gs2007} for a discussion of the connections between plan recognition and natural language processing.) \citet{azn1998,aznb1997} seek to recognise what ``quest'' a player is pursuing in an online adventure game, for which they use a dynamic Bayesian network \citep{dk1989} whose parameters are learned using a corpus of historical play data. Similarly, \citet{g2010} trains an Input-Output Hidden Markov Model \citep{bf1995} to predict a player's goal in an action-adventure game. Closely related is the work of \citet{ba2004,ba2003}, who compute goal probabilities as a product of conditional action probabilities which are learned using a corpus of observed plan executions. This work was later extended to recognise hierarchical sub-goals \citep{ba2006}.

		\subsection{\bf Recursive Reasoning} \label{sec:recreas}

Autonomous agents often base their decisions on explicit beliefs about the state of the environment and, possibly, the mental states of other agents. The mental states of other agents may, in turn, also contain beliefs about the environment and mental states of other agents. This nesting of beliefs leads to a possibly infinite reasoning process of the form ``I believe that you believe that I believe...''. While the modelling methods discussed in the previous sections do not model such nested beliefs, methods of \emph{recursive reasoning} use explicit representations of nested beliefs and ``simulate'' the reasoning processes of other agents to predict their actions.

Game theorists first addressed infinitely nested beliefs in the context of incomplete information games, in which some components of the game (such as players' payoff functions) are not common knowledge \citep{h1962}. In Bayesian games \citep{h1967}, an early precursor of type-based reasoning (see Section~\ref{sec:typebased}), the infinite regress is resolved by assuming that the private elements of players are drawn from a distribution that is common knowledge. While this assumption allows for an elegant equilibrium analysis \citep{h1968b}, creating such a setting is rather impractical when designing an autonomous agent that is interacting with unknown other agents. Recursive reasoning methods follow a more direct approach by \emph{approximating} the belief nesting down to a fixed recursion depth. As a prototypical example, assume agent A is modelling another agent B. In order to choose an action, A predicts the next action of B by simulating the decision making of B given what A believes about B. This requires a prediction of A's next action from B's perspective, given what A believes B to believe about A, and so on. The recursion is terminated at some predetermined depth by fixing the action prediction to some probability distribution, e.g. uniform probabilities. The prediction at the bottom of this recursion is passed up to the above recursion level to choose an optimal action at that level, which in turn is passed to the next higher level, and so on, until agent A can make its \emph{actual} choice at the beginning of the recursion. Note that the recursion assumes that each agent believes to have more sophisticated (deeper) beliefs than the other agent. Another central assumption is that each agent assumes the other agent to be \emph{rational}\footnote{We already saw instances of this rationality assumption in utility reconstruction (Section~\ref{sec:policyrec-utility}) and some approaches for plan recognition (Section~\ref{sec:planrec}).} in that it will choose optimal actions with respect to its beliefs.

The method proposed by \citet{cm1996search} implements the recursion outlined above for game tree search in games with alternating moves. Here, an agent model specifies the agent's evaluation function for game states as well as the evaluation function the agent believes its opponent to use, and so on. As the authors point out, the well-known minimax algorithm for zero-sum games \citep{cm1983} is a special case of this method in which the evaluation function of the opponent is simply the negative of one's own function. The ``Recursive Modeling Method'' (RMM) \citep{gd2000,gd1995,gdw1991} also implements the above recursion, with the added complexity that agents may be uncertain about the exact model of other agents, such as their payoff function and recursion depth. In the above example, agent A has additional probabilistic beliefs about the possible models of agent B. During the recursion, A has to predict B's action under each possible model, adding an extra branching factor to the recursion. The resulting predictions are then weighted by the probabilities in A's beliefs about B's models. \citet{gnk1998} also show how these beliefs can be updated after new observations, which involves the recursive updating of the beliefs of other agents, such that A updates its own belief about B's models, and B's expected belief about A's possible models, and so on. \citet{vd1995} show how the recursion in RMM can be made more efficient by pruning branches in the recursion tree which are expected to have no or minimal influence on the final choice of the agent.

RMM is the precursor of the Interactive POMDP (I-POMDP) \citep{gd2005}. In a POMDP \citep{s1971}, an agent makes sequential decisions based on its belief about the state of the environment, which is represented as a probability distribution over possible states and updated based on incomplete and uncertain observations. I-POMDPs modify POMDPs by adding model spaces to the environment state, such that an agent has beliefs about the environment state \emph{and} the models of other agents. Agent models are categorised into ``sub-intentional'' and ``intentional'' models. A sub-intentional model defines any non-recursive mapping from observation histories to action probabilities, such as the finite state automata used in the work of \citet{pg2017}. In contrast, intentional models are themselves defined as I-POMDPs with beliefs about the environment and models of other agents. I-POMDPs are solved via a finite recursion as outlined above: To choose an optimal action, agent A has to solve the I-POMDP of agent B for each of its intentional models, which in turn requires solving the I-POMDP of agent A for each model ascribed to A by B, and so on, down to some fixed recursion depth. At the bottom of the recursion are standard POMDPs in which other agents are treated as ``noise'' in the transition and observation dynamics. These POMDPs can be solved directly using existing methods \citep{klc1998} and their solutions are passed up the recursion tree. Several exact and approximate solution methods for I-POMDPs have been proposed, including methods based on model equivalence \citep{rdg2006}, particle filtering \citep{dg2009}, value iteration \citep{dp2008}, policy iteration \citep{sd2015}, and structural problem reduction \citep{hl2013}. \citet{nbmw2012} propose an even more complex modification of I-POMDPs in which agents are also uncertain about the transition and observation models of the environment.

An alternative to quantitative (probabilistic) representations of uncertainty (as used in RMM and I-POMDPs) are qualitative belief representations based on logics, such as dynamic epistemic logic (DEL) \citep{ba2011,lpw2010}. Epistemic logics are characterised by a knowledge operator $K_i \phi$ (or $B_i \phi$) which expresses that agent $i$ ``knows'' (or ``believes'') the formula $\phi$. For example, $K_i K_j K_i \phi$ corresponds to ``agent $i$ knows that agent $j$ knows that agent $i$ knows $\phi$''. The semantics of $K_i \phi$ are defined such that it holds true if $\phi$ is true in all world states that agent $i$ believes the world may be in. The dynamic aspect of DEL is given by event operators (actions) that can modify ontic and epistemic facts in the world via pre/post-conditions, similar to other planning languages such as STRIPS \citep{fn1971}. Several planning methods have been proposed that use such epistemic logics. \citet{mbfm2015} and \citet{kg2015} both propose methods that solve epistemic planning problems using classical planning algorithms. \citet{vdh2002} solve epistemic planning problems using model checking algorithms. \citet{gll2007} propose a framework based on the situation calculus \citep{mh1969} for reasoning about beliefs and coordination in agent teams.

\begin{table}[t]
	\btab
		\tabrow{dvb2017}{yes}{no}{yes}{yes}{no}{\al}{discrete}{full}
		\tabrow{kg2015}{no}{no}{yes}{yes}{no}{\si}{discrete}{partial}
		\tabrow{mbfm2015}{no}{no}{yes}{yes}{no}{\si}{discrete}{partial}
		\tabrow{sd2015}{yes}{yes}{yes}{yes}{no}{\si}{discrete}{partial}
		\tabrow{dvv2013}{yes}{no}{yes}{yes}{no}{\si}{discrete}{full}
		\tabrow{hl2013}{yes}{yes}{yes}{yes}{no}{\si}{discrete}{partial}
		\tabrow{nbmw2012}{yes}{yes}{yes}{yes}{no}{\si}{discrete}{partial}
		\tabrow{ba2011}{no}{no}{yes}{yes}{no}{\si}{discrete}{partial}
		\tabrow{lpw2010}{no}{no}{yes}{yes}{no}{\si}{discrete}{partial}
		\tabrow{dg2009}{yes}{yes}{yes}{yes}{no}{\si}{discrete}{partial}
		\tabrow{dp2008}{yes}{yes}{yes}{yes}{no}{\si}{discrete}{partial}
		\tabrow{gll2007}{no}{no}{yes}{yes}{yes}{\al}{discrete}{partial/full}
		\tabrow{rdg2006}{yes}{yes}{yes}{yes}{no}{\si}{discrete}{partial}
		\tabrow{gd2005}{yes}{yes}{yes}{yes}{no}{\si}{discrete}{partial}
		\tabrow{ctj2004}{yes}{no}{yes}{yes}{no}{\si}{discrete}{full}
		\tabrow{vdh2002}{no}{no}{yes}{yes}{no}{\si}{discrete}{partial}
		\tabrow{gd2000,gd1995}{yes}{no}{yes}{yes}{no}{--\as}{discrete}{--\ass}
		\tabrow{gnk1998}{yes}{no}{yes}{yes}{no}{--\as}{discrete}{partial}
		\tabrow{cm1996search}{no}{no}{yes}{yes}{no}{\al}{discrete}{full}
		\tabrow{vd1995}{yes}{no}{yes}{yes}{no}{--\as}{discrete}{--\ass}
		\tabrow{gdw1991}{yes}{no}{yes}{yes}{no}{--\as}{discrete}{--\ass}
	\etab
	\caption{Assumptions in papers for recursive reasoning methods. \as No explicit move order defined. \ass No explicit observation model used.}
	\label{tab:recreas}
\end{table}

Given the belief nesting, an important question is how deep the recursion should be to achieve a robust interaction with humans and other agents. This question has been addressed extensively by researchers in behavioural game theory and experimental psychology \citep{chc2015,gdy2012,wl2010,ydf2008,ctj2004,hz2002}. For example, \citet{ctj2004} develop a simple recursive reasoning model in which an agent at recursion level $k$ has probabilistic beliefs regarding what level $k' < k$ the other agent uses. The beliefs are assumed to be correct, in that they are derived from a population distribution over recursion depths which is represented as a Poisson distribution. After ``fitting'' the model based on a large corpus of human play data, the authors find that humans reason on average at depth 1.5, i.e. one or two levels down the recursion. In addition to experiments with humans, some research pitted artificial recursive reasoning agents against each other to see what reasoning depths are most useful. For example, \citet{dvb2017,dvv2013} test their specific agents in domains such as repeated rock-paper-scissors and sequential negotiation, and find that reasoning levels deeper than 2 do not provide significant benefits in their setting.

		\subsection{\bf Graphical Models} \label{sec:graphmod}

The modelling methods discussed in the previous sections are based on rather abstract formulations of multiagent systems, in which much of the system's structure is left implicit. For example, a common formulation describes an environment which at any time is in some abstract state $s$, and transition probabilities between states are specified by some function $T(s,a,s')$ where $a$ is a tuple containing the agents' actions. In addition, an agent's utility is commonly defined as a general function $u(s,a)$ that depends on the state and joint action. What is left implicit in such formulations are the precise relations between the state components $s = (s_1,...,s_m)$ (e.g. some components may depend on other components); how state components interact with the agents' decisions $a = (a_1,...,a_n)$ (e.g. some agents may disregard certain components in their decisions); and the precise dependencies of utilities on state components and actions (e.g. an agent's utility may depend on the actions of some agents but not on others).

\emph{Graphical models} make such dependencies explicit by using graph representations of multiagent systems. The advantage of making this structure explicit is that, if the interaction is only over a short horizon,\footnote{Graphical models can represent sequential interactions by adding additional nodes for each time step in the interaction, as well as dependencies between nodes in different time steps \citep{jn2011}. Unfortunately, this approach does not scale efficiently with the number of time steps \citep[e.g.][]{dzc2009,gp2003ws}.} it can lead to compact models and more efficient algorithms, similarly to how Bayesian networks exploit conditional independence relations for compactness and efficient inference \citep{kf2009,p1988}. Moreover, graphical models can be used as detailed mental models of how other agents may view the interaction.

The basic building block of many graphical models is the ``Influence Diagram'' (ID) \citep{hm2005,hm1984}. An ID is a graphical representation of a single-agent decision problem. IDs use three types of nodes: chance nodes, which describe the components in the environment state; decision nodes, whose values the agent has to choose; and utility nodes, which determine the agent's utilities. Directed edges between nodes indicate dependence relations, e.g. the parent nodes of a decision node constitute the information that is used by the agent for that particular decision. A solution to an ID is a set of optimal decision rules, one for each decision node, which specify action probabilities for each input to the decision nodes \citep{s1986}. Given a set of decision rules, an ID can be reduced to a normal Bayesian network by replacing each decision node with a chance node whose conditional probabilities are specified by the corresponding decision rule. One can then use standard inference algorithms \citep{p1988} to compute a variety of queries, such as expected utilities and the probability of certain events. The ``Multi-Agent Influence Diagram'' (MAID) \citep{kb2003} extends IDs by assigning each decision and utility node to one of several agents. Graphical games \citep{vk2002,kls2001,l2000} can be viewed as a special type of MAID that have only decision and utility nodes. These works on MAID and graphical games show how the graph structure can be exploited for efficient computation of Nash equilibrium solutions \citep{n1950}.

Graphical models, such as IDs and MAIDs, can be used by an agent to model the decision making and \emph{domain conceptualisation} of other agents. For example, an existing parent relation between a chance node $X$ and a decision node $D$ encodes the belief that the modelled agent incorporates $X$ in its decision for $D$; conversely, the absence of such a relation encodes the belief that the modelled agent does not account for $X$ in its decision for $D$ (or not directly). Several works have used graphical models for such mental representations of other agents. \citet{sg1999} use IDs to model the capabilities, beliefs, and preferences of other agents. They show how the parameters of an ID may be modified to reflect the observed behaviour of an agent, focusing on learning the agent's preferences by modifying the utility nodes in the ID. \citet{nj2004} also propose methods to learn the utility function in an ID for an observed agent. They relax the usual rationality assumption, which requires that the agent choose actions to strictly optimise its utilities, by allowing for random deviations from optimality. \citet{mk2000} define a probabilistic epistemic logic (cf. Section~\ref{sec:recreas}) to represent and infer the beliefs of agents, and use IDs to derive an agent's decision rules given its inferred beliefs and assuming the agent is rational. \citet{cabl2013} use conditional preference (CP) networks \citep{bbdh2004} to model the preferences of players based on their negotiation dialogues. The resulting CP-nets are used to predict the players' actions by computing an equilibrium solution over the preferences encoded by the CP-nets. \citet{ckp2000} use IDs to represent the preferences of patients in a clinical trial and propose an algorithm for effective preference elicitation, which is the problem of deciding what questions to ask patients to obtain additional information about their preferences.

\begin{table}[t]
	\btab
		\tabrow{cabl2013}{yes}{no}{yes}{yes}{no}{\al}{discrete}{partial/full}
		\tabrow{zd2012}{yes}{yes}{yes}{yes}{no}{\si}{discrete}{partial}
		\tabrow{dcz2010,dzc2009}{yes}{yes}{yes}{yes}{no}{\si}{discrete}{partial}
		\tabrow{gp2008}{yes}{yes}{yes}{yes}{no}{\si}{discrete}{partial}
		\tabrow{nj2004}{yes}{yes}{yes}{yes}{no}{--\asss}{discrete}{full}
		\tabrow{gp2003}{yes}{yes}{yes}{yes}{no}{\si}{discrete}{partial}
		\tabrow{kb2003}{yes}{no\as}{yes}{yes}{no}{\si}{discrete}{partial/--\as}
		\tabrow{vk2002}{yes}{no\as}{yes}{yes}{no}{\si}{--\ass/disc.}{--\ass/--\as}
		\tabrow{kls2001}{yes}{no\as}{yes}{yes}{no}{\si}{--\ass/disc.}{--\ass/--\as}
		\tabrow{l2000}{yes}{no\as}{yes}{yes}{no}{\si}{--\ass/disc.}{--\ass/--\as}
		\tabrow{mk2000}{yes}{no}{yes}{yes}{no}{--\asss}{discrete}{partial}
		\tabrow{ckp2000}{yes}{no}{yes}{yes}{no}{\al}{mixed}{partial/full}
		\tabrow{sg1999}{yes}{yes}{yes}{yes}{no}{\si}{discrete}{full}
	\etab
	\caption{Assumptions in papers for graphical methods. \as Does not model repeated interactions. \ass Does not model environment states. \asss Does not define move order.}
	\label{tab:graphmod}
\end{table}

Graphical models can also represent uncertainty over multiple hypothesised models of other agents (as in type-based reasoning; see Section~\ref{sec:typebased}) and nested beliefs (as in recursive reasoning; see Section~\ref{sec:recreas}). ``Networks of Influence Diagrams'' (NIDs) \citep{gp2008,gp2003} achieve this as follows: A NID is a single-rooted graphical model in which each node is a MAID. The root node of a NID represents the perspective of the modelling agent, and directed edges $A \rightarrow_{j,D} B$ indicate that the agent whose view is represented by the MAID in node $A$ believes that agent $j$ uses the MAID in node $B$ to make some decision $D$. If multiple such edges exist for the same agent $j$ and decision $D$, then the MAID in $A$ may contain a new chance node specifying the probabilistic belief of the modelling agent for each edge. The MAID in node $B$ may contain beliefs about other agents, and cycles in a NID are used to represent recursive reasoning. NIDs are solved by first solving the leaves of the NID, which are normal MAIDs that can be solved with existing methods \citep{kb2003}. The solutions are decision rules for the decision nodes, which are passed to the parents in the NID, transforming them into MAIDs that can be solved, and so forth. A related model is the ``Interactive Dynamic ID'' (I-DID) \citep{dzc2009} which was designed as a graphical representation of I-POMDPs \citep{gd2005} (cf. Section~\ref{sec:recreas}). In contrast to NIDs, which compute equilibrium solutions for a set of agents, I-DIDs are designed for subjective decision making of a single agent in a system containing multiple agents. This means that I-DIDs do not represent the decisions of other agents as decision nodes (as in MAIDs) but rather as chance nodes whose conditional probabilities are governed by the possible models ascribed to the agents, which may themselves be I-DIDs. Models and uncertainties over models are represented in a new ``model node''. I-DIDs represent temporal relations between nodes by ``unrolling'' the network for each time step in the interaction, such that edges between nodes in successive time steps indicate temporal dependencies (similar to dynamic Bayesian networks; \citet{dk1989}). To manage the exponential growth of possible agent models after new observations, methods have been proposed which cluster behaviourally similar models \citep{zd2012,dcz2010,dzc2009}.

		\subsection{\bf Group Modelling} \label{sec:groupmod}

Most methods surveyed in the earlier sections use models that make predictions about a \emph{single} agent, following the agent model shown in Figure~\ref{fig:model}. For methods that predict an agent's actions, such as policy reconstruction (Section~\ref{sec:policyrec}), type-based reasoning (Section~\ref{sec:typebased}), and recursive reasoning (Section~\ref{sec:recreas}), modelling single agents is predicated on the assumption that agents choose actions independently from each other, as defined in Section~\ref{sec:assum}. Thus, many papers proceed by explaining their methods for a single agent, with the underlying idea that the same method can be used to maintain separate models for each other agent. Note that this separation does not mean that agents ignore each other, since the models may base their predictions on the observed actions of other agents. Nonetheless, there are important cases in which it may be preferable to use \emph{group models} which make joint predictions about a group of agents.

One such case is when agents have significant randomisation and \emph{correlation} in their action choices (cf. Section~\ref{sec:assum}), which cannot be captured by independent models. An example of this case is the concept of correlated equilibrium \citep{a1974}, which generalises the Nash equilibrium by defining the equilibrium as a joint distribution over agents' actions rather than independent distributions. Many of the existing methods for policy reconstruction and type-based reasoning can be used to learn such action correlations, essentially by combining all other agents into a single agent whose action space is the Cartesian product of the agents' actions. This approach allows a model to capture action correlations by making predictions about the joint probability of actions. However, this approach may scale poorly since the action space of the ``combined agent'' grows exponentially in the number of combined agents and actions. A middle-path is to partition the other agents into smaller groups such that there is high expected correlation within groups but only little or no correlation between groups \citep[an approach commonly used in probabilistic state estimation, e.g.][]{ar2016jair,bk1998}. The modelling agent can then use separate group models for each group.

Even when there is no significant randomisation in action choices, group models can often be more efficient and accurate by exploiting additional structure in the group. In particular, agent groups may act as \emph{teams} which utilise structure such as roles within teams, dynamic formation of subteams, ``divide-and-conquer'' division of goals into sub-goals, as well as predefined joint plans and communication protocols \citep{sv1999aij,t1997,gk1996,cl1991}. Knowledge of such structure can be used by group models to effectively limit the search space. For example, the behaviours of agents in a coordinated team, when observed in isolation, may not be very informative (and even possibly misleading) as to the intended goals of the agents. However, when the same behaviours are interpreted in the context of a team, they may give important clues as to the goal and plan of the team \citep{t1996}. In this spirit, a number of methods have been proposed which model teams rather than individual agents.

Section~\ref{sec:classif} already surveyed several works which use classification methods to identify teams and team strategies \citep{bdfe2010,lsma2009,ilsk2008,ss2007,s2004,rv2000}. In addition, methods have been developed which model the physical formation and movement patterns of teams. \citet{ev2011} use a hierarchical clustering method to extract clusters of similar movement trajectories from log data in the small size multi-robot league of RoboCup. During a game, the method observes an incomplete trajectory from the opponent team and classifies it into one of the extracted clusters, which allows it to predict future movements and compute counter-strategies. \citep{rv2001} propose a method for simulated robot soccer which uses a predefined set of opponent models that specify probabilities of field positions for each player in the opponent team, given their initial positions and ball movements. Starting with a prior distribution over models, Bayesian updates are performed after new movement observations and the most probable model is used in the planning stage. \citet{lmvh2005} also consider simulated robot soccer and use unsupervised symbolic learning to extract movement patterns from observations. \citet{kks2006} propose a method for the RoboCup simulated coach competition which can classify ``patterns'' (defined as exploitable weaknesses in an opponent team's strategy) by extracting feature vectors that include formation statistics, and comparing them to previously learned models from log data.

While the above methods learn and use models of \emph{opponent} teams, an agent may also need to model its \emph{own} team. This is important in problems of ad hoc (or impromptu) teamwork \citep{skkr2010,bm2005}, in which an agent has to collaborate ``on the fly'' with an established but previously unknown team, without opportunities for prior coordination with the team members. \citet{bm2005} consider such a setting in the context of robot soccer, in which the team uses ``plays'' from a set of predefined plays, called the playbook. Each play specifies roles for the agents in the team along with sequences of synchronised actions for each role, as well as applicability and termination conditions for the play. A pickup player joins an established team but is not informed about the currently used plays nor its role in the plays. Assuming that the pickup player has access to a playbook, its task is to find the correct plays and its role within the plays. One proposed method to achieve this task is to compute a matching score for each play based on how well the play matches the observed actions in the team, and to select the play that has the highest matching score. \citet{bs2015} consider a similar setting in the Half-Field Offense domain \citep{hfo2016} and use reinforcement learning to learn optimal collaboration policies for the pickup player in a range of previously encountered teams. During a new game, the pickup player uses the optimal policy for the past team which is most similar to the new team. Bayesian probabilities are calculated to quantify similarity between past teams and the new team, using models of past teams which predict transition probabilities between observed game states.

\begin{table}[t]
	\btab
		\tabrow{bs2015}{yes}{no}{yes}{no}{yes}{\si}{contin./disc.}{full/--\as}
		\tabrow{zyk2012}{no}{no}{yes}{no}{no}{\si}{discrete}{partial}
		\tabrow{bk2011}{no}{no}{yes}{no}{no}{\si}{--\ass/discrete}{partial}
		\tabrow{ev2011}{yes}{no}{yes}{yes}{no}{\si}{contin.}{full/--\as}
		\tabrow{zl2011}{no}{no}{yes}{no}{no}{\si}{--\ass/discrete}{partial}
		\tabrow{bkl2010}{no}{no}{yes}{no}{no}{\si}{--\ass/discrete}{full}
		\tabrow{ss2008}{no}{no}{yes}{no}{no}{\si}{--\ass/discrete}{partial}
		\tabrow{kks2006}{yes}{no}{yes}{no}{no}{\si}{contin.}{full/--\as}
		\tabrow{bm2005}{no}{yes}{yes}{no}{yes}{\si}{contin.}{full/--\as}
		\tabrow{lmvh2005}{yes}{no}{yes}{no}{no}{\si}{contin.}{full/--\as}
		\tabrow{sm2004}{yes}{no}{yes}{mixed\asss}{yes}{\si}{discrete}{partial}
		\tabrow{kpt2002}{yes}{no}{yes}{mixed}{yes}{\si}{--\ass/discrete}{partial}
		\tabrow{rv2001}{yes}{no}{yes}{yes}{no}{\si}{discrete}{full/--\as}
		\tabrow{t1996}{no}{no}{yes}{no}{no}{\si}{mixed/disc.}{full}
	\etab
	\caption{Assumptions in papers for group modelling. \as Actions are not directly observed. \ass Does not define environment states. \asss Actions may be correlated in team (joint) policies and independent in the lower (individual) policies.}
	\label{tab:groupmod}
\end{table}

In addition to a large body of work on plan recognition for single agents (cf. Section~\ref{sec:planrec}), there is a growing body of work on multiagent plan recognition in which the modelling agent attempts to infer the goals and plans of an entire team of agents. Thus, plan libraries specify team plans that utilise additional structure such as roles within teams and division into subteams. \citet{t1996} extends a previous method \citep[][cf. Section~\ref{sec:planrec-hierarch}]{tr1995} by using a hierarchical team plan library. Teams can be divided into subteams which must be assigned to exactly one role in the team. Similar to the original method, the new method quickly commits to a single plan hypothesis and repairs inconsistencies via backtracking in the plan hierarchy. \citet{sm2004} propose an extension of the abstract hidden Markov model \citep[][cf. Section~\ref{sec:planrec-hierarch}]{bvw2002} in which top-level joint policies for the team select lower-level policies for individual agents which are executed in a decentralised way. The proposed method proceeds similarly to the original work by defining the plan inference based on dynamic Bayesian networks and using particle filtering to perform the inference. \citet{ss2008} use a hierarchical plan library specified with AND/OR trees similar to the model of \citet{gg2009} (cf. Section~\ref{sec:planrec-hierarch}), with extra elements to specify the number of agents needed to commence a plan and special nodes in plan trees to generate and resolve subteams. The authors show how this additional structure can be utilised to prune the search space in the recognition task. \citet{kpt2002} propose a method which infers a team's current plan based on overheard communications between team members, using plan and team hierarchies. \citet{bkl2010} show NP-completeness in a restricted version of multiagent plan recognition, in which team plans are defined as matrices that specify a sequence of synchronised actions for a subset of agents. This work was subsequently extended to allow for interleaved plan execution and incomplete observation traces \citep{bk2011}. \citet{zl2011} consider a similar formulation to \citet{bkl2010} but allow for partial observations. The proposed method frames the plan recognition problem as a satisfiability problem by automatically generating a set of constraints from the plan library and observations, which are solved using a MAX-SAT solver. In later work, \citet{zyk2012} propose a similar SAT-based recognition approach using action specifications in the STRIPS planning language rather than matrix-based plan libraries.

		\subsection{\bf Other Relevant Methods} \label{sec:other}

In this section we briefly discuss several other relevant methods, namely implicit modelling, hypothesis testing for agent models, and safe best-response methods.

			\subsubsection{Implicit Modelling} \label{sec:other-implicit}

This survey focused on \emph{explicit} modelling of other agents, in which agent models implement the mapping shown in Figure~\ref{fig:model}. In contrast, \emph{implicit} modelling does not produce explicit models of other agents, but implicitly encodes aspects of other agents (such as their behaviours) in other structures or reasoning processes. For example, ``expert'' algorithms, which learn to follow the best expert policy from a given set of such policies \citep[e.g.][]{c2014,fm2004}, can be viewed as implicit modelling in that each expert policy may be optimal against a particular opponent and, thus, implicitly encode the opponent's behaviour without making explicit predictions about that opponent. Implicit modelling based on expert algorithms has been shown to be effective in variants of Poker \citep{bjbb2013,hshb2005}. Other examples of implicit modelling include learning logical action descriptions in the context of other agents \citep{igos2010,ges2004}; modelling other agents as part of the MDP transition dynamics \citep{hzt2017}; and using opponent features in a neural network to learn expected action utilities \citep{hbkd2016}. A potential advantage of implicit modelling is that it may more naturally exploit synergies between modelling and planning by merging the two processes. Advantages of explicit modelling are that the models are decoupled from the planning and may thus be used by different planning algorithms, and that explicit models are more amenable to direct inspection. It is also possible to combine these two forms of modelling, e.g. \citet{acr2015mipc} combine expert algorithms with type-based reasoning (cf. Section~\ref{sec:typebased}).

			\subsubsection{Hypothesis Testing for Agent Models} \label{sec:other-testing}

Agent models may make incorrect or inaccurate predictions. This is one of the main motivations of type-based reasoning methods (Section~\ref{sec:typebased}), which consider a set of alternative models and compute Bayesian posteriors to find the most accurate model. However, such Bayesian methods generally cannot tell us about the \emph{correctness} of models, since the posteriors quantify a relative likelihood of models but not absolute truth. Thus, even if all probability points to one model, that model may still be almost arbitrarily incorrect in that it merely has to support the observations, i.e. assign non-zero probabilities. An alternative approach is to view a model as a \emph{hypothesis} and to decide, based on the observations, whether or not to reject the model. For agent models that predict actions, this question can be decided using methods for statistical hypothesis testing. For example, agents have been proposed which maintain models of action frequencies of other agents and conduct hypothesis tests over these models by comparing their predicted action probabilities with the average action frequencies over some window of past actions \citep{cs2014,cs2007,fy2003}. \citet{ar2015} propose an efficient sampling-based algorithm which uses ``score functions'' to compute test statistics from observations and learns the test distribution during the interaction, based on which a frequentist hypothesis test is performed. Given such methods, if an agent persistently rejects a model, it may decide to change the model (e.g. by using a different learning method) or to resort to some kind of default policy such as a minimax strategy \citep{nm1944}.

\begin{table}[t]
	\btab
		\tabrow{hzt2017}{yes}{yes}{yes}{yes}{no}{\si}{discrete}{full}
		\tabrow{hbkd2016}{yes}{no\asss}{yes}{no}{no}{\si}{mixed}{full}
		\tabrow{ar2015}{yes}{yes}{no}{yes}{no}{--\as}{mixed/disc.}{full}
		\tabrow{acr2015mipc}{yes}{no}{yes}{yes}{no}{\si}{discrete}{full}
		\tabrow{bjbb2013}{yes}{no}{no}{yes}{no}{\al}{discrete}{partial/full}
		\tabrow{wbmp2011}{yes}{yes}{yes}{yes}{no}{\si}{discrete}{full}
		\tabrow{igos2010}{no}{no}{yes}{yes}{no}{\si}{mixed/disc.}{full}
		\tabrow{jb2009}{yes}{no}{no}{yes}{no}{\al}{discrete}{partial/full}
		\tabrow{jzb2008}{yes}{no}{no}{yes}{no}{\al}{discrete}{partial/full}
		\tabrow{cs2007}{yes}{yes}{yes}{yes}{no}{\si}{--\ass/discrete}{full}
		\tabrow{hshb2005}{yes}{no}{no}{yes}{no}{\al}{discrete}{partial/full}
		\tabrow{mr2005}{no}{no}{no}{yes}{no}{\si}{discrete}{full}
		\tabrow{mb2004}{yes}{yes}{no}{yes}{no}{\si}{--\ass/discrete}{full}
		\tabrow{ges2004}{no}{yes}{no}{yes}{no}{--\as}{discrete}{full}
		\tabrow{fy2003}{yes}{yes}{no}{yes}{no}{\si}{--\ass/discrete}{full}
		\tabrow{srv2000}{yes}{no}{no}{yes}{no}{\si}{mixed/disc.}{partial}
		\tabrow{cm1996search93}{no}{no}{yes}{yes}{no}{\al}{discrete}{full}
	\etab
	\caption{Assumptions in papers for other relevant methods. \as Does not define move order. \ass Does not define environment states. \asss Modelled agent may change behaviour between episodes but not during episode.}
	\label{tab:other}
\end{table}

			\subsubsection{Using Models Safely} \label{sec:other-safety}

An agent can utilise models of other agents by incorporating the models' predictions into the agent's planning process. For example, if a model predicts the actions of another agent, then these predictions can be used directly by a planner to evaluate different courses of actions, resulting in an action policy that is strictly optimised with respect to the model. A potential problem with this approach is that the computed policy may be exploitable by other agents if the used agent models are inaccurate. To address this issue, several methods have been proposed which compute ``safe'' (or ``robust'') best-response policies to models. These methods often use a parameter of the form $\delta \in [0,1]$ which regulates a tradeoff between safety and exploitability, such that one extreme corresponds to strict optimisation with respect to the agent models (optimal if models correct, but exploitable otherwise) and the other extreme corresponds to choosing a safe policy which may not achieve optimal performance but is less exploitable (e.g. minimax). For example, \citet{wbmp2011} model an opponent as a space of models in the proximity of the empirical frequency model, with distance bounded by $\delta$, and compute a best-response against the worst-case model from this space. Other examples of safe/robust best-response methods include the works of \citet{jb2009,jzb2008,mb2004,cm1996search93}. A related idea is the use of ``ideal'' agent models \citep{srv2000}. For example, \citet{mr2005} propose to learn the weaknesses of an opponent, which are defined as states in which the opponent deviates from some ideal ``teacher'' policy.

	\section{Open Problems} \label{sec:openprob}

We conclude our survey by discussing nine open problems which we believe have not been sufficiently addressed in the literature and may provide fruitful avenues of future research.

		\subsection{Synergistic Combination of Modelling Methods}

This survey has outlined a landscape of methodologies, each with their individual purposes, strengths, and weaknesses. An interesting and relatively unexplored question is how these methods might be combined to complement their strengths and weaknesses. As an example, type-based reasoning methods have been combined with policy reconstruction methods, where the former allow for fast initial adaptation while the latter generate new types during the interaction \citep{ar2013,bsk2011}. These examples use a modular combination, by encapsulating the policy reconstruction methods into a special kind of type. In the long-term, an important question is whether we can find a single representation and approach that can naturally generate various modelling capabilities, including the ones discussed in this survey, such that the modelling processes synergistically inform one another. We believe there is much ground for fertile research investigating such combinations and approaches.

		\subsection{Policy Reconstruction under Partial Observability}

Many domains are characterised by partial observability, in which agents receive incomplete and uncertain observations about the environment and the actions of other agents (cf. Section~\ref{sec:assum}). The existence of partial observability can make the modelling task significantly more difficult, since a modelling agent now has to take into account the possibility of incorrect and/or missing information. Different symbolic and probabilistic approaches have been proposed to deal with partial observability, especially in methods for classification, plan recognition, recursive/epistemic reasoning, and graphical models. However, as can be seen in Table~\ref{tab:policyrec}, relatively little work exists on the problem of learning models of agent behaviours (i.e. policy reconstruction) under partial observability conditions, with most efforts focusing on extensive form games with incomplete information (e.g.~Poker). Moreover, existing methods often assume that observation probabilities can be derived via provided domain knowledge \citep[e.g.][]{pg2017,sbl2005}. Thus, additional research is needed for the development of methods which can effectively reconstruct behaviour models under partial observability, and methods which can deal with partial observability in the absence of domain knowledge.

		\subsection{Safe and Efficient Model Exploration}

Agents that model other agents can consider the possibility of taking actions so as to explore certain aspects of the other agents' behaviours, and in the process gain new information which may lead to better model predictions. However, such actions may carry a risk in that they may modify the behaviour of the modelled agents in unintended ways. Although the importance of safe model exploration was recognised almost 20 years ago \citep{cm1998}, it has since received relatively little attention in the community.\footnote{Indeed, the vast majority of current plan recognition methods assume that the modelling agent does not interact at all with the modelled agents (cf. Section~\ref{sec:planrec}).} Current solutions are based on look-ahead exploration to estimate the value of information of available actions \citep{acr2016aij,cb2003,cm1999}. However, the exponential complexity of such methods makes them intractable in complex settings, indicating the need for new, more efficient approaches for safe model exploration. Closely related areas are active learning \citep{b2012}, preference elicitation \citep{b2002,ckp2000}, and Bayesian experimental design \citep{cv1995}. However, these problems usually assume that the cost of experiments/queries and their possible outcomes are known beforehand, while in our case the (long-term) cost of exploratory actions are initially unknown and there may be no crisp definition of ``outcomes''.

		\subsection{Efficient Discovery of Decision Factors}

Closely related to safe model exploration, it remains a significant open question how to efficiently and effectively discover the relevant factors in an agent's decision making (cf. Section~\ref{sec:assum}). Current methods either assume that this knowledge is given, include all possible decision factors in the model, or engage in an exhaustive combinatorial search to identify the relevant factors (cf. Section~\ref{sec:policyrec-conditional}). However, these approaches are bound to be intractable or inefficient in complex, realistic applications that involve large numbers of decision factors (such as long interaction histories and high-dimensional state descriptions). Hence, more research is needed to develop methods which can efficiently discover the relevant decision factors in an agent's decision making.

		\subsection{Computationally Efficient Implementations}

Modelling methods are part of a larger agent architecture which may include many other elements, such as modules for perception (e.g. vision, natural language), communication, and planning. In domains such as commercial video games, the system will in addition have to graphically render the game world and simulate its physics \citep{m2009}. All of these additional elements can be computationally expensive. As a result, the task of modelling other agents will usually be allocated only a small fraction of the available computational resources. Therefore, to be useful in practice, modelling methods need highly efficient implementations, similar to other recent applications \citep{alphago2016,bbjt2015}. Efficient implementations may include the use of efficient data structures, parallel computing architectures, and iterative model updates which process only new observations rather than re-processing past observations. Such implementation issues have received relatively little attention in the literature, thus additional research is needed to develop efficient implementations.

		\subsection{Modelling Changing Behaviours}

A common assumption still found in many modelling methods is that the modelled agent, in particular its behaviour, will not change during the course of the interaction (cf. Section~\ref{sec:assum}). However, such an assumption is easily violated in applications in which other agents may learn and adapt, and especially in interactions with humans. Modelling changing behaviours is notoriously difficult due to the essentially unconstrained nature of what other agents may do in the future. Some methods attempt to address this issue by allowing for varying degrees of changing behaviours, such as that behaviours must converge in the limit \citep{cs2007}, that agents may switch periodically between different stationary behaviours \citep{hzt2017,bb2007}, by defining behaviours as blackbox mappings over the entire interaction history \citep{acr2016aij}, or by prioritising recent observations over past ones \citep{ar2013,bdsb2004}. Still, many methods are unable to deal with changing behaviours, especially methods for classification, plan recognition, and recursive reasoning. Hence, the design of methods which can effectively learn to identify, track, and predict changing behaviours remains a significant open problem, one which will be a crucial element in the quest for full autonomy.

		\subsection{Modelling with Action Duration}

The vast majority of surveyed methods (with the exception of some plan recognition methods; cf. Section~\ref{sec:planrec}) assume that actions have instant effects, meaning that actions are completed immediately after they are taken. Even in domains such as robot soccer, where actions such as passing the ball from one player to another have durations, current modelling methods work at a level of abstraction that renders such actions as though they have instant effects \citep[e.g.][]{bdfe2010,kfcv2002}. It is not clear if existing modelling methods require non-trivial modification to handle actions with durations, or if this can be addressed sufficiently via such action abstractions. In fact, it is unclear if the notion of action duration may be better viewed as an issue of activity recognition, which is the task of inferring action labels from state data and usually takes place at a lower abstraction level than the modelling methods surveyed in this article (cf. Section~\ref{sec:assum}). Given that many realistic applications involve actions with durations, we believe that such questions will require further research and clarification.

		\subsection{Modelling in Open Multiagent Systems}

Virtually all of the surveyed works in this article assume \emph{closed} multiagent systems, in which the number of agents in the system remains constant throughout the interaction, and all agents begin the interaction at the same time. This is in contrast to \emph{open} multiagent systems, in which agents may enter and leave the system at any time during the interaction, without necessarily notifying other agents. Many important applications are characterised by such openness, such as ad-hoc wireless networks \citep{rt1999} and web-based systems for collaborative computing \citep{mmrn2014}. In addition, a fully autonomous agent engaged in lifelong learning \citep{hr2013} may itself enter and leave many multiagent systems. While some works investigated modelling other agents in open multiagent systems \citep{ceds016,hjs2006,rw2003}, it remains a significant open challenge to develop efficient modelling methods for such systems. Transfer learning, which is the process of reusing past experiences to improve learning in new tasks, could be a useful element in such methods \citep[e.g.][]{bskr2013}.

		\subsection{Autonomous Model Contemplation and Revision}

While the methods discussed in this survey enable an autonomous agent to reason about other agents in highly sophisticated ways, they do not generally tell the agent if the used methods are the right ones in any given setting. As a result, it is possible that the agent may use inadequate and possibly misleading models of other agents, without ever realising it. For example, learning-based methods for policy reconstruction are usually restricted by the structure of the model (e.g. decision trees, finite state automata) but do not tell the modelling agent if the model structure is even capable of capturing an agent's behaviour. Type-based reasoning can utilise a space of models, but the Bayesian beliefs do not generally tell an agent if the model space is sufficient. Methods for plan recognition that use plan libraries suffer from essentially the same limitation (cf. Section~\ref{sec:planrec-planning}). To detect such insufficiencies, a modelling agent requires the ability to introspectively reason about the adequacy and correctness of its modelling processes, and ultimately the ability to autonomously revise its model structures and modelling processes. Statistical hypothesis testing can be used to reason about the incorrectness of models (cf. Section~\ref{sec:other-testing}), but such methods do not tell us \emph{why} a model is incorrect and \emph{how} it may be revised. In fact, it is likely that the conventional notion of correctness is too strict, and that different notions of adequacy (such as the degree to which a model allows the modelling agent to complete its task) may be needed. The current generation of intelligent agents fall short of full autonomy in part because they lack the ability to contemplate such questions, and we believe there is much research to be done to address these issues.


	\section{Conclusion} \label{sec:conclusion}

This survey identified seven major methodologies for agents modelling other agents. Surveyed methods include policy reconstruction, which seeks to reconstruct an agent's decision making based on its observed actions; type-based reasoning, which maintains beliefs over a space of alternative decision-making models to identify the most likely models based on observed actions; classification methods, which use machine learning to predict additional properties of interest such as behaviour classes and agent identities; plan recognition, which seeks to identify an agent's goals and plans using hierarchical action descriptions or domain models; recursive reasoning, which predicts an agent's actions by modelling its beliefs and the beliefs it ascribes to other agents; graphical models, which utilise graph structures to represent detailed dependence relations in an agent's decision making; and group modelling, in which models make joint predictions about a group of agents rather than single agents. We also covered other relevant methods, including implicit modelling, hypothesis testing for agent models, and safe best-response methods. Finally, we identified a number of open problems which can provide fertile grounds for future research. Our survey of the literature shows that there exists a very large body of work on the topic of agents modelling other agents, broadly addressing questions of algorithmic design, experimental evaluation, theoretical guarantees, computational complexity, and observational constraints. As research in artificial intelligence continues to pursue the goal of creating autonomous agents that interact with other agents to accomplish tasks in complex dynamic domains, we expect to see continued development towards addressing these questions. Our hope is that this survey will contribute to this continued development by summarising the current state of research and exposing important open problems.

	\section*{Acknowledgements}

This survey benefited from comments and suggestions of many colleagues, which we would like to thank here: Michael Rovatsos, Nolan Bard, Michael Littman, Karl Tuyls, Christopher Geib, Subramanian Ramamoorthy, Alex Lascarides, Gal Kaminka, and three anonymous reviewers. This work took place in the Learning Agents Research Group (LARG) at The University of Texas at Austin. LARG research is supported in part by grants from the National Science Foundation (IIS-1637736, IIS-1651089, IIS-1724157), Intel, Raytheon, and Lockheed Martin. Stefano Albrecht is supported by a Feodor Lynen Research Fellowship from the Alexander von Humboldt Foundation. Peter Stone serves on the Board of Directors of Cogitai, Inc. The terms of this arrangement have been reviewed and approved by The University of Texas at Austin in accordance with its policy on objectivity in research.

	\appendix
	\section[\hspace{5em}Clarification for Assumption Tables]{Clarification for Assumption Tables} \label{sec:appendix}

Tables \ref{tab:policyrec}--\ref{tab:other} list assumptions for each surveyed paper in the corresponding sections. Assumptions are in the order in which they are discussed in Section~\ref{sec:assum}. The first five assumptions concern the agents to be modelled and include:

\begin{description}[labelindent=10pt,itemsep=1pt]
	\item (1) whether they make stochastic or deterministic action choices
	\item (2) whether they have changing or non-changing behaviours
	\item (3) whether their relevant decision factors are a priori known
	\item (4) whether they make independent or correlated action choices
	\item (5) whether they have common or conflicting goals
\end{description}

The last three assumptions concern the environment within which the interaction takes place and include:

\begin{description}[labelindent=10pt,itemsep=1pt]
	\item (6) the order in which agents take actions (simultaneous, alternating)
	\item (7) the representation used for environment states and actions (discrete, continuous, mixed)
	\item (8) the observability of environment states and actions (full, partial)
\end{description}

For assumptions (7) and (8), we may distinguish between states and actions by using a ``state/action'' notation. Additional comments are provided in the table captions.

We note that while many works state all or most of the above assumptions explicitly, there are also many works which are rather vague about some assumptions. In vague cases, we tried to infer assumptions based on our understanding of the provided descriptions.


	\section*{References}
	\bibliographystyle{elsarticle-harv} 
	\bibliography{survey_aij}

\end{document}